\PassOptionsToPackage{expansion=false}{microtype}
\documentclass[manuscript]{acmart}

\AtBeginDocument{%
  }

\setcopyright{acmlicensed}
\copyrightyear{2026}
\acmYear{2026}
\acmDOI{XXXXXXX.XXXXXXX}
\acmConference[FAccT '26]{ACM Conference on Fairness, Accountability, and Transparency}{June 2026}{Montreal, Canada}
\acmISBN{978-1-4503-XXXX-X/2026/06}

\usepackage{booktabs}
\usepackage[expansion=false]{microtype}
\usepackage{amsmath}
\usepackage{graphicx}
\usepackage{tikz}
\usepackage{pgfplots}
\pgfplotsset{compat=1.18}
\usetikzlibrary{arrows.meta,positioning,shapes.geometric,calc}
\begin{document}

\title{Why Policy Prompts Are Not Enough: Learning Decision Playbooks for Fair and Auditable LLM Governance}

\author{Sanket Badhe}
\email{sanketbadhe@google.com}
\affiliation{%
  \institution{Google}
  \city{Mountain View}
  \state{California}
  \country{USA}
}

\author{Priyanka Tiwari}
\affiliation{%
  \institution{Purdue University}
  \country{USA}
}

\renewcommand{\shortauthors}{Anonymous Author(s)}

\begin{abstract}
Large language models (LLMs) are widely used to enforce written policies in content moderation, legal compliance, and safety governance. In practice, practitioners place the official policy document directly into the model's \emph{Policy Prompt} and run a frozen \emph{Evaluation Model} as a zero-shot classifier. However, high-level policy text leaves real-world edge cases, dialect nuances, and statutory exceptions underspecified, leading to low accuracy and high false-positive rates on protected subgroups and compliant edge cases. We introduce \textsc{LEDGER}, an automated framework that keeps the official Policy Prompt frozen while learning a compact, auditable \emph{Decision Playbook} from training errors. Within each policy section, \textsc{LEDGER} organizes rules into a two-tier decision list that checks safe-harbor exceptions before violation criteria, iteratively refining rules so they fix boundary errors without overturning predictions the base policy already got right. Across HateXplain, Social Bias Frames (SBIC.v2), and LexGLUE Unfair-ToS, learned Decision Playbooks improve single-pass Macro-F1 by $4.2$ to $11.1$ points over the base Policy Prompt while reducing subgroup false-positive rates by up to $24.1$ percentage points, and transfer zero-shot across unseen Evaluation Models.
\end{abstract}




\maketitle

\section{Introduction}
\label{sec:intro}

Organizations in trust and safety, legal compliance, healthcare administration, and AI safety increasingly use large language models (LLMs) to enforce written policies \cite{gorwa2020algorithmic,bai2022constitutional,raji2020closing,inan2023llamaguard,han2024wildguard}. A standard practice is to place the organization's official policy document (such as a platform speech policy, a statutory compliance checklist, or a safety guideline) directly into the prompt of a fast, frozen \emph{Evaluation Model} (\emph{Policy Prompt}) to classify incoming items in a single forward pass. Keeping both the model weights and the official policy text frozen has clear practical advantages: the policy document remains the single authoritative source of truth, inference cost is low, and practitioners do not need to retrain model weights when policies change \cite{vaccaro2020contesting,raji2020closing}.

However, a frozen Policy Prompt alone is often insufficient for accurate and fair enforcement. Written policies state general rules at a high level of abstraction. They rarely specify how to resolve recurring edge cases, such as when a contextual exception overrides a general prohibition, how to interpret domain vernacular or quoted material, or where to draw the boundary between adjacent violation categories. When an Evaluation Model applies a high-level policy prompt zero-shot, it often relies on surface keyword associations learned during pretraining. As a result, it misclassifies ambiguous boundary cases and produces high false-positive rates on posts mentioning minority demographic groups \cite{sap2019risk,davidson2019racial,dixon2018measuring,borkan2019nuanced}.

To solve this problem, we introduce \textsc{LEDGER}, an automated method that keeps both the official Policy Prompt and the Evaluation Model frozen while learning a compact, auditable \emph{Decision Playbook} from a small labeled sample ($240$ to $480$ examples). Instead of adding a flat list of rules where exception rules and violation rules compete on the same keywords, \textsc{LEDGER} organizes rules within each policy section into two ordered tiers: \textbf{Tier~1 exception rules} (which identify benign or lower-severity edge cases) are checked first, before \textbf{Tier~2 violation rules} (which define when a policy violation applies). Every rule also includes an explicit boundary stating when it does not apply. During training, the Evaluation Model cites the rule ID it used for each prediction. \textsc{LEDGER} uses these citations to track which baseline errors each rule fixes and which previously correct predictions it breaks, allowing it to keep accurate rules, revise rules that cause new errors or subgroup false positives, and remove unused rules.

In summary, we make three main contributions:
\begin{itemize}
    \item \textbf{Two-Tier Decision Playbooks:} We show why unstructured lists of corrective rules overturn predictions that the base policy already got right, and introduce a two-tier playbook structure that checks exception rules before violation rules within each policy section (Sections~\ref{sec:problem} and \ref{sec:method}.1).
    \item \textbf{Rule-Level Auditing and Refinement (\textsc{LEDGER}):} We develop \textsc{LEDGER}, an iterative algorithm that uses per-prediction rule citations to track each rule's fixes and regressions on training data, locking reliable rules, refining rules that cause errors or subgroup false positives, and pruning unused rules (Sections~\ref{sec:method}.2--\ref{sec:method}.4).
    \item \textbf{Evaluation Across Moderation and Legal Compliance:} We evaluate \textsc{LEDGER} on three public benchmarks using their official policy definitions: HateXplain \cite{mathew2021hatexplain}, Social Bias Frames (SBIC.v2) \cite{sap2020social}, and LexGLUE Unfair-ToS \cite{lippi2019claudette,chalkidis2022lexglue} (Sections~\ref{sec:setup}--\ref{sec:results}). Across all three tasks, \textsc{LEDGER} improves single-pass Macro-F1 by $4.2$ to $11.1$ points over the base Policy Prompt, lowers subgroup false-positive rates by up to $24.1$ percentage points, and transfers zero-shot to unseen Evaluation Models.
\end{itemize}

\section{Related Work}
\label{sec:related}

\subsection{Policy Governance, Safeguard Classifiers, and Algorithmic Auditing}
Across regulatory compliance, clinical guidelines, and platform governance, written policies state general principles that require operational rules when applied to individual cases \cite{gorwa2020algorithmic,gordon2022jury}. \citet{vaccaro2020contesting} showed that users contesting automated moderation decisions need consistent explanations tied to specific policy rules. Similarly, \citet{raji2020closing} argued that algorithmic auditing requires documentation linking specific system components to their empirical impact across subgroups before deployment. In LLM safety and alignment, Constitutional AI \cite{bai2022constitutional} and dedicated safeguard classifiers such as Llama Guard \cite{inan2023llamaguard} and WildGuard \cite{han2024wildguard} enforce safety taxonomies through weight fine-tuning or high-level normative prompts, but leave the model to resolve boundary conflicts implicitly. \textsc{LEDGER} complements these approaches by keeping both the Evaluation Model and the official Policy Prompt frozen while learning a versioned, section-indexed Decision Playbook in which each rule is paired with its four-cell training audit record and cited at inference time.

\subsection{Subgroup Bias and False Positives in Policy Enforcement}
Classifiers that enforce policies in sensitive domains often treat surface mentions of protected groups, minority dialects, or standard legal disclaimers as indicators of policy violations \cite{waseem2016hateful,davidson2017automated,founta2018large,lippi2019claudette}. In hate speech and toxicity detection, \citet{sap2019risk} and \citet{davidson2019racial} showed that classifiers frequently misclassify African American English (AAE) and reclaimed in-group terms as hate speech. \citet{dixon2018measuring} and \citet{borkan2019nuanced} further showed that benign mentions of identity terms such as \emph{gay}, \emph{Muslim}, or \emph{Jewish} increase false-positive rates on those subgroups \cite{lees2022perspective}. Benchmarks such as HateXplain \cite{mathew2021hatexplain} and Social Bias Frames (SBIC.v2) \cite{sap2020social} were created to measure both classification accuracy and subgroup fairness across explicit and implied bias \cite{elsherief2021latent,hartvigsen2022toxigen,vidgen2021learning}. Similar boundary errors occur in legal compliance under EU Directive 93/13/EEC (CLAUDETTE / Unfair-ToS in LexGLUE) \cite{lippi2019claudette,chalkidis2022lexglue}, where classifiers confuse lawful scope-of-service descriptions or bilateral termination clauses with unlawful liability waivers or unilateral modifications. Prior debiasing methods typically rely on fine-tuning model weights, reweighting training data, or supervising attention with human rationales \cite{mathew2021hatexplain,devlin2019bert}. Our approach reduces subgroup false positives without updating model weights by learning explicit Tier~1 safe-harbor exceptions with bounded exclusion conditions.

\subsection{Automatic Prompt Optimization, Rule Induction, and Decision Lists}
Automatic prompt optimization methods iteratively refine natural-language instructions using labeled validation examples. APE \cite{zhou2023large} and OPRO \cite{yang2024large} generate whole-prompt candidates based on overall validation scores. ProTeGi \cite{pryzant2023automatic} and TextGrad \cite{yuksekgonul2025optimizing} generate textual critiques of training errors to rewrite prompts, while DSPy/MIPRO \cite{khattab2024dspy,opsahlong2024optimizing} jointly optimize instructions and demonstrations. A related line of work uses LLMs to induce discrete sets of rules or hypotheses: Hypotheses-to-Theories (HtT) \cite{zhu2024large} induces reasoning rules and filters them by empirical accuracy, HypoGeniC \cite{zhou2024hypothesis} maintains a bank of data-driven hypotheses scored on training examples, and ExpeL \cite{zhao2024expel} extracts natural-language insights from contrasting trajectories through add, edit, and removal operations. Classical knowledge-acquisition systems such as Ripple-Down Rules (RDR) \cite{compton1990philosophical}, decision lists \cite{rivest1987learning}, and certifiably optimal rule lists (CORELS) \cite{angelino2017learning} similarly showed that ordered if-then-else rules and ``except-when'' refinements prevent newly added rules from corrupting existing rules. \textsc{LEDGER} adapts these exception-list principles to single-pass LLM policy enforcement. Unlike flat LLM rule banks \cite{zhu2024large,zhou2024hypothesis,zhao2024expel} or whole-prompt rewriters \cite{pryzant2023automatic,yuksekgonul2025optimizing}, \textsc{LEDGER} organizes rules by policy section into two ordered tiers (checking Tier~1 exception rules before Tier~2 violation rules), includes an explicit boundary on when each rule does not apply, and uses cited rule IDs (\texttt{rule\_ids}) to track which training examples each rule fixes or breaks under subgroup fairness constraints.

\section{Problem Formulation and Risk Decomposition}
\label{sec:problem}

\subsection{Multi-Section Policy Enforcement with a Frozen LLM}
Consider a policy enforcement task over an input space $\mathcal{X}$ (such as user posts, contract clauses, or support tickets) and a discrete set of $C$ policy decisions $\mathcal{Y} = \{y_1, \dots, y_C\}$. We are given an official \emph{Policy Prompt} $P_0$ divided into ordered policy sections $\mathcal{S} = (s_1, \dots, s_M)$ that define the criteria for assigning inputs to decisions in $\mathcal{Y}$, with a default compliant outcome $y_{\text{default}} = y_{s_M} \in \mathcal{Y}$. Each example $(x_i, y_i) \in \mathcal{X} \times \mathcal{Y}$ may also be associated with protected demographic or topical subgroup attributes $G_i \subseteq \mathcal{G}$ (with indicator $g_i = \mathbb{I}[G_i \neq \emptyset]$), for which false-positive predictions into a primary violation class $y_{\star} = y_{s_1} \in \mathcal{Y}$ (or any non-compliant class $\mathcal{Y} \setminus \{y_{\text{default}}\}$) should be controlled. Specifically, on any dataset $\mathcal{D}$, we define the \emph{subgroup severe-violation false-positive rate} $\text{FPR}_{\text{SG}}(\mathcal{B}; \mathcal{D})$ and the \emph{subgroup benign false-positive rate} $\text{FPR}_{\text{SG-Benign}}(\mathcal{B}; \mathcal{D})$ as:
\begin{align}
    \text{FPR}_{\text{SG}}(\mathcal{B}; \mathcal{D}) &= \frac{\bigl|\bigl\{ i \in \mathcal{D} : G_i \neq \emptyset \;\wedge\; y_i \neq y_{s_1} \;\wedge\; \hat{y}(x_i; \mathcal{B}) = y_{s_1} \bigr\}\bigr|}{\bigl|\bigl\{ i \in \mathcal{D} : G_i \neq \emptyset \;\wedge\; y_i \neq y_{s_1} \bigr\}\bigr|}, \label{eq:fpr_sg} \\
    \text{FPR}_{\text{SG-Benign}}(\mathcal{B}; \mathcal{D}) &= \frac{\bigl|\bigl\{ i \in \mathcal{D} : G_i \neq \emptyset \;\wedge\; y_i = y_{\text{default}} \;\wedge\; \hat{y}(x_i; \mathcal{B}) \neq y_{\text{default}} \bigr\}\bigr|}{\bigl|\bigl\{ i \in \mathcal{D} : G_i \neq \emptyset \;\wedge\; y_i = y_{\text{default}} \bigr\}\bigr|}, \label{eq:fpr_benign}
\end{align}
with the per-subgroup rate $\text{FPR}_{\text{SG}}(g; \mathcal{B}, \mathcal{D})$ defined analogously by restricting the numerator and denominator of Equation~\eqref{eq:fpr_sg} to examples with $g \in G_i$.

At deployment, a frozen \emph{Evaluation Model} $\mathcal{M}_{\text{eval}}$ classifies each input $x \in \mathcal{X}$ in a single pass while keeping both its parameters and the official policy text $P_0$ fixed. Using a small labeled training set $\mathcal{D}_{\text{train}}$ and validation set $\mathcal{D}_{\text{val}}$, an offline \emph{Optimizer Model} $\mathcal{M}_{\text{opt}}$ learns a structured \emph{Decision Playbook} $\mathcal{B} = (c_1, \dots, c_K)$ containing $K \le K_{\max}$ operational clauses indexed by policy section $s \in \mathcal{S}$. At test time, $\mathcal{M}_{\text{opt}}$ is not used. Instead, $\mathcal{M}_{\text{eval}}$ takes $(P_0, \mathcal{B}, x)$ and outputs a structured JSON prediction containing a brief rationale, the cited clause identifiers $R(x; \mathcal{B}) \subseteq \{c_1, \dots, c_K\}$ (\texttt{applied\_rule\_ids}), and the decision $\hat{y}(x; \mathcal{B}) \in \mathcal{Y}$.

We formulate Decision Playbook learning as a constrained optimization problem. Our goal is to learn a playbook $\mathcal{B} \in \mathcal{H}$ that maximizes classification Macro-F1 while keeping the subgroup false-positive rate $\text{FPR}_{\text{SG}}(\mathcal{B})$ within a tolerance $\delta_{\text{FPR}} \ge 0$ of the base policy's rate, and ensuring that overall accuracy does not drop below the base policy $P_0$ ($\mathcal{B} = \emptyset$):
\begin{equation}
    \max_{\mathcal{B} \in \mathcal{H}} \; \text{MacroF1}(\mathcal{B}) \quad \text{subject to} \quad \text{FPR}_{\text{SG}}(\mathcal{B}) \le \text{FPR}_{\text{SG}}(\emptyset) + \delta_{\text{FPR}}, \quad \text{Acc}(\mathcal{B}) \ge \text{Acc}(\emptyset).
    \label{eq:constrained_obj}
\end{equation}

\subsection{The Natural Preservation Set and the Rule-Overlap Trap}
To see why standard rule induction struggles to satisfy Equation~\eqref{eq:constrained_obj}, let $\hat{y}_i^{(0)} = \hat{y}(x_i; \emptyset)$ denote the baseline prediction of $\mathcal{M}_{\text{eval}}$ using the frozen Policy Prompt $P_0$ alone. The baseline Policy Prompt splits any dataset $\mathcal{D}$ into two disjoint subsets: the \emph{baseline error set} $\mathcal{E}_{0}(\mathcal{D}) = \{i \in \mathcal{D} : \hat{y}_i^{(0)} \neq y_i\}$ and the \emph{natural preservation set} $\mathcal{P}_{0}(\mathcal{D}) = \{i \in \mathcal{D} : \hat{y}_i^{(0)} = y_i\}$, which contains all examples the base policy already classifies correctly. When a Decision Playbook $\mathcal{B}$ is added, each example $i \in \mathcal{D}$ falls into one of four mutually exclusive sets:
\begin{align}
    \text{Fix}(\mathcal{B}; \mathcal{D}) &= \bigl\{ i \in \mathcal{E}_{0}(\mathcal{D}) : \hat{y}_i^{(\mathcal{B})} = y_i \bigr\}, &
    \text{Preserve}(\mathcal{B}; \mathcal{D}) &= \bigl\{ i \in \mathcal{P}_{0}(\mathcal{D}) : \hat{y}_i^{(\mathcal{B})} = y_i \bigr\}, \label{eq:fix} \\
    \text{Regress}(\mathcal{B}; \mathcal{D}) &= \bigl\{ i \in \mathcal{P}_{0}(\mathcal{D}) : \hat{y}_i^{(\mathcal{B})} \neq y_i \bigr\}, &
    \text{Miss}(\mathcal{B}; \mathcal{D}) &= \bigl\{ i \in \mathcal{E}_{0}(\mathcal{D}) : \hat{y}_i^{(\mathcal{B})} \neq y_i \bigr\}. \label{eq:miss}
\end{align}
Because $\mathcal{P}_{0}(\mathcal{D}) = \text{Preserve}(\mathcal{B}; \mathcal{D}) \cup \text{Regress}(\mathcal{B}; \mathcal{D})$, the net accuracy change of any playbook $\mathcal{B}$ relative to the base policy equals the number of baseline errors it fixes in $\mathcal{E}_{0}(\mathcal{D})$ minus the number of regressions it causes on the natural preservation set $\mathcal{P}_{0}(\mathcal{D})$:
\begin{equation}
    \Delta \text{Acc}(\mathcal{B}; \mathcal{D}) = \text{Acc}(\mathcal{B}; \mathcal{D}) - \text{Acc}(\emptyset; \mathcal{D}) = \frac{|\text{Fix}(\mathcal{B}; \mathcal{D})| - |\text{Regress}(\mathcal{B}; \mathcal{D})|}{|\mathcal{D}|}.
    \label{eq:net_acc}
\end{equation}
When each prediction cites its active clause(s) $c_k \in R(x_i; \mathcal{B})$ with citation set $\mathcal{C}(c_k) = \{i \in \mathcal{D} : c_k \in R(x_i; \mathcal{B})\}$, Equation~\eqref{eq:net_acc} can be approximated as a sum of per-clause net contributions:
\begin{equation}
    \Delta \text{Acc}(\mathcal{B}; \mathcal{D}) \approx \frac{1}{|\mathcal{D}|} \sum_{k=1}^{K} \underbrace{\Bigl( |\text{Fix}(c_k; \mathcal{D})| - |\text{Regress}(c_k; \mathcal{D})| \Bigr)}_{\text{NetGain}(c_k; \mathcal{D})},
    \label{eq:clause_decomp}
\end{equation}
with exact equality when each example cites a single clause ($|R(x_i; \mathcal{B})| = 1$) and uncited examples keep $\hat{y}_i^{(0)}$. When $\mathcal{M}_{\text{eval}}$ cites multiple clauses ($|R(x_i; \mathcal{B})| > 1$) or cites no clause ($R(x_i; \mathcal{B}) = \emptyset$), \textsc{LEDGER} records both unweighted set membership $i \in \mathcal{C}(c_k)$ (so every clause cited on a regression is flagged for review) and fractional counts $1/|R(x_i; \mathcal{B})|$, while logging uncited regressions in $\mathcal{U}_{\text{reg}}$ and uncited errors in $\mathcal{U}_{\text{err}}$. We also report the \emph{preservation retention rate} $|\text{Preserve}(\mathcal{B}; \mathcal{D})| / |\mathcal{P}_{0}(\mathcal{D})|$, which measures the fraction of initially correct predictions preserved by $\mathcal{B}$.

Equation~\eqref{eq:clause_decomp} shows why unstructured rule lists fall into the \emph{Rule-Overlap Trap}. Naive rule induction inspects only the baseline errors $\mathcal{E}_{0}(\mathcal{D}_{\text{train}})$ and proposes positive rules of the form $c_k : \psi_k^{+}(x) \Rightarrow y_k$. When these rules are not tested against the natural preservation set $\mathcal{P}_{0}(\mathcal{D}_{\text{train}})$, a trigger condition $\psi_k^{+}(x)$ written to fix errors in $\mathcal{E}_{0}$ often matches examples in $\mathcal{P}_{0}$ that share the same keywords but belong to another category, increasing $|\text{Regress}(c_k)|$. Moreover, if an input matches both an exception rule and a violation rule, an unordered rule list leaves the decision ambiguous. Avoiding these regressions requires ordering exception rules before violation rules, adding explicit exclusion boundaries, and verifying each rule on $\mathcal{P}_{0}(\mathcal{D}_{\text{train}})$.

\section{Learning Decision Playbooks with LEDGER}
\label{sec:method}

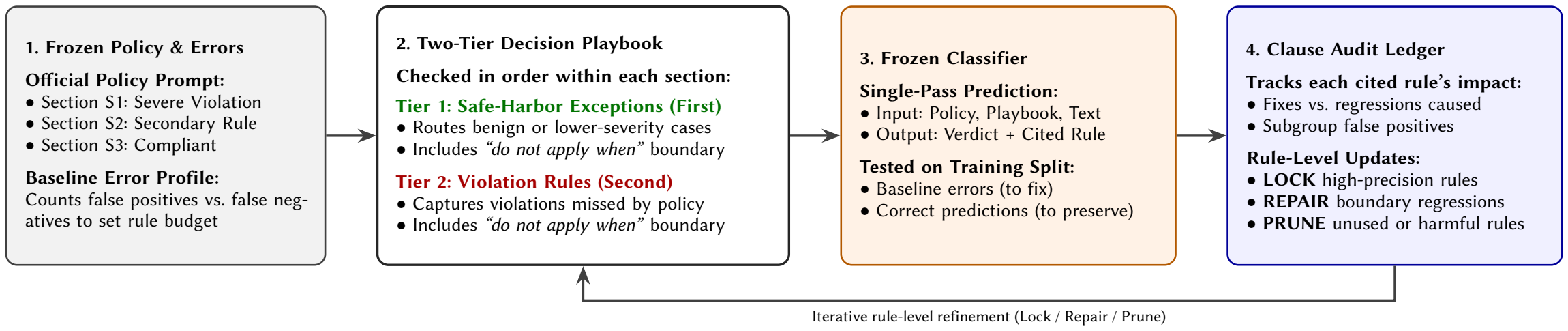
\begin{figure*}[t]
\centering
\resizebox{\textwidth}{!}{%
\begin{tikzpicture}[
    font=\sffamily\footnotesize,
    node distance=0.65cm,
    box/.style={draw=black!75, rounded corners=4pt, align=left, inner sep=7pt, line width=0.65pt, minimum height=3.35cm},
    arrow/.style={-{Stealth[length=2.4mm, width=1.8mm]}, line width=0.85pt, draw=black!75}
]
% Box 1: Policy & Error Profile
\node[box, fill=gray!10, text width=3.65cm] (policy) {
    \textbf{1. Frozen Policy \& Errors}\\[4pt]
    \textbf{Official Policy Prompt:}\\
    $\bullet$ Section S1: Severe Violation\\
    $\bullet$ Section S2: Secondary Rule\\
    $\bullet$ Section S3: Compliant\\[4pt]
    \textbf{Baseline Error Profile:}\\
    Counts false positives vs.\ false negatives to set rule budget
};

% Box 2: Two-Tier Decision Playbook
\node[box, fill=white, draw=black!85, line width=0.85pt, text width=4.85cm, right=of policy] (playbook) {
    \textbf{2. Two-Tier Decision Playbook}\\[4pt]
    \textbf{Checked in order within each section:}\\[3pt]
    \textcolor{green!45!black}{\textbf{Tier 1: Safe-Harbor Exceptions (First)}}\\
    $\bullet$ Routes benign or lower-severity cases\\
    $\bullet$ Includes \emph{``do not apply when''} boundary\\[4pt]
    \textcolor{red!65!black}{\textbf{Tier 2: Violation Rules (Second)}}\\
    $\bullet$ Captures violations missed by policy\\
    $\bullet$ Includes \emph{``do not apply when''} boundary
};

% Box 3: Frozen Evaluation Model
\node[box, fill=orange!10, draw=orange!70!black, text width=3.85cm, right=of playbook] (eval) {
    \textbf{3. Frozen Classifier}\\[4pt]
    \textbf{Single-Pass Prediction:}\\
    $\bullet$ Input: Policy, Playbook, Text\\
    $\bullet$ Output: Verdict + Cited Rule\\[4pt]
    \textbf{Tested on Training Split:}\\
    $\bullet$ Baseline errors (to fix)\\
    $\bullet$ Correct predictions (to preserve)
};

% Box 4: Clause Audit Ledger
\node[box, fill=blue!6, draw=blue!60!black, text width=3.85cm, right=of eval] (ledger) {
    \textbf{4. Clause Audit Ledger}\\[4pt]
    \textbf{Tracks each cited rule's impact:}\\
    $\bullet$ Fixes vs.\ regressions caused\\
    $\bullet$ Subgroup false positives\\[4pt]
    \textbf{Rule-Level Updates:}\\
    $\bullet$ \textbf{LOCK} high-precision rules\\
    $\bullet$ \textbf{REPAIR} boundary regressions\\
    $\bullet$ \textbf{PRUNE} unused or harmful rules
};

% Clean center-aligned forward arrows
\draw[arrow] (policy.east) -- (playbook.west);
\draw[arrow] (playbook.east) -- (eval.west);
\draw[arrow] (eval.east) -- (ledger.west);

% Clean bottom feedback loop from Audit Ledger back to Playbook
\draw[arrow] (ledger.south) -- ++(0,-0.45cm) -| node[pos=0.25, below, font=\sffamily\scriptsize] {Iterative rule-level refinement (Lock / Repair / Prune)} (playbook.south);
\end{tikzpicture}%
}
\caption{Overview of \textsc{LEDGER}. Starting from a frozen multi-section Policy Prompt and its baseline training error profile (1), \textsc{LEDGER} organizes rules within each section into a two-tier Decision Playbook that checks Tier~1 safe-harbor exceptions before Tier~2 violation rules, each with an explicit \emph{``do not apply when''} boundary (2). A frozen Evaluation Model classifies inputs in a single pass and cites the active rule ID (3). The Clause Audit Ledger tracks each rule's fixes, regressions, and subgroup false positives to \textbf{LOCK} reliable rules, \textbf{REPAIR} boundary regressions, and \textbf{PRUNE} inactive rules across rounds (4).}
\label{fig:architecture}
\end{figure*}

Figure~\ref{fig:architecture} illustrates \textsc{LEDGER}. The method applies to any multi-section Policy Prompt $P_0 = (s_1, \dots, s_M)$ by combining a two-tier decision-list structure with per-clause verification on the natural preservation set $\mathcal{P}_{0}(\mathcal{D}_{\text{train}})$.

\subsection{Two-Tier Decision Lists with Exclusion Boundaries}
\textsc{LEDGER} represents the playbook $\mathcal{B}$ as a \emph{two-tier decision list} $\mathcal{H}_{\text{DL}}$ indexed by policy section. Each clause $c_k \in \mathcal{B}$ is a tuple $(\texttt{id}_k, s_k, \tau_k, y_k, \psi_k^{+}, \psi_k^{-})$, where $s_k \in \{s_1, \dots, s_{M-1}\}$ is the policy section, $\tau_k \in \{\mathcal{T}_1, \mathcal{T}_2\}$ is the precedence tier within section $s_k$, $y_k \in \mathcal{Y}$ is the target decision, $\psi_k^{+}: \mathcal{X} \to \{0,1\}$ is a positive trigger condition, and $\psi_k^{-}: \mathcal{X} \to \{0,1\}$ is an exclusion condition (``do not apply when $\psi_k^{-}(x)$ holds''). Clause $c_k$ fires on an input $x$ if and only if the positive trigger holds and the exclusion condition does not:
\begin{equation}
    A(c_k) = \bigl\{ x \in \mathcal{X} : \psi_k^{+}(x) = 1 \;\wedge\; \psi_k^{-}(x) = 0 \bigr\}.
    \label{eq:set_diff}
\end{equation}
The two conditions in Equation~\eqref{eq:set_diff} separately control the two terms of $\text{NetGain}(c_k)$ in Equation~\eqref{eq:clause_decomp}. The positive trigger $\psi_k^{+}(x)$ covers baseline errors in $\mathcal{E}_{0}$ (increasing $|\text{Fix}(c_k)|$), while the exclusion condition $\psi_k^{-}(x)$ prevents $A(c_k)$ from matching examples in the natural preservation set $\mathcal{P}_{0}$ (limiting $|\text{Regress}(c_k)|$). To generalize beyond the training set, $\mathcal{M}_{\text{opt}}$ writes $\psi_k^{+}$ and $\psi_k^{-}$ in terms of structural or pragmatic properties (such as speaker intent, target type, or contractual exception criteria) rather than specific training keywords.

To avoid conflicts when multiple rules match an input, clauses in $\mathcal{H}_{\text{DL}}$ are evaluated in a fixed decision-list order:
\begin{equation}
    \mathcal{T}_1^{(s_1)} \succ \mathcal{T}_2^{(s_1)} \succ \mathcal{T}_1^{(s_2)} \succ \mathcal{T}_2^{(s_2)} \succ \dots \succ \mathcal{T}_1^{(s_{M-1})} \succ \mathcal{T}_2^{(s_{M-1})} \succ P_0.
    \label{eq:precedence_order}
\end{equation}
Within each policy section $s_m$ (ordered from $s_1$ to $s_{M-1}$):
\begin{enumerate}
    \item \textbf{Tier~1 Exception Rules / Safe Harbors ($\mathcal{T}_1^{(s_m)}$):} Checked first whenever an input $x$ potentially falls under section $s_m$. Tier~1 clauses identify contextual exceptions and route the input to the compliant class $y_{\text{default}}$ or a lower-severity section $y_{s_2}$ (provided the exclusion boundary $\psi_k^{-}(x)$ is not triggered). Within $\mathcal{T}_1^{(s_m)}$, clauses targeting $y_{\text{default}}$ are listed before clauses targeting $y_{s_2}$, and ties are broken by alphanumeric clause ID (\texttt{S1.B1}, \texttt{S1.B2}, \dots).
    \item \textbf{Tier~2 Violation Rules ($\mathcal{T}_2^{(s_m)}$):} Checked in alphanumeric ID order (\texttt{S1.C1}, \texttt{S1.C2}, \dots) only if no Tier~1 rule in $\mathcal{T}_1^{(s_m)}$ applied. Tier~2 clauses specify the criteria for assigning $x$ to the section violation label $y_{s_m}$. If no clause applies across sections $s_1, \dots, s_{M-1}$, $\mathcal{M}_{\text{eval}}$ defaults to the base Policy Prompt $P_0$.
\end{enumerate}

\subsection{Initializing Rule Capacity from Baseline Error Residuals}
Different policies and datasets fail in different ways under the base policy $P_0$: some suffer mainly from false-positive over-enforcement on benign edge cases, whereas others suffer from false-negative under-enforcement on subtle violations. Before generating initial candidate rules for section $s_m$, \textsc{LEDGER} counts the baseline confusion entries $N_{y \to \hat{y}}^{(0)} = |\{i \in \mathcal{D}_{\text{train}} : y_i = y \wedge \hat{y}_i^{(0)} = \hat{y}\}|$ and computes the section's over-enforcement residual $E_{\text{over}}^{(s_m)}$ and under-enforcement residual $E_{\text{under}}^{(s_m)}$:
\begin{equation}
    E_{\text{over}}^{(s_m)} = \sum_{y \neq y_{s_m}} N_{y \to y_{s_m}}^{(0)}, \qquad
    E_{\text{under}}^{(s_m)} = \sum_{\hat{y} \neq y_{s_m}} N_{y_{s_m} \to \hat{y}}^{(0)}.
    \label{eq:flow}
\end{equation}
In Round~1, \textsc{LEDGER} allocates the clause budget $K_{\max}$ between Tier~1 and Tier~2 in proportion to Equation~\eqref{eq:flow}. When over-enforcement dominates ($E_{\text{over}}^{(s_1)} \ge E_{\text{under}}^{(s_1)}$, as in HateXplain and Unfair-ToS), more capacity is assigned to Tier~1 exception rules to filter out false positives, using true violations from $\mathcal{P}_{0}(\mathcal{D}_{\text{train}})$ to define their exclusion boundaries $\psi_k^{-}$. Conversely, when under-enforcement dominates ($E_{\text{under}}^{(s_1)} > E_{\text{over}}^{(s_1)}$, as in SBIC.v2), Tier~1 is kept compact and more capacity is assigned to Tier~2 violation rules to capture missed violations. Because clauses are partitioned by policy section $s_m$, \textsc{LEDGER} evaluates both the full candidate playbooks proposed by $\mathcal{M}_{\text{opt}}$ and a section-wise coordinate combination formed by pairing the highest-precision clause block for each section on $\mathcal{D}_{\text{train}}$.

\subsection{Per-Clause Credit Assignment and Lock-Repair-Prune Updates}
Standard prompt optimizers \cite{zhou2023large,yang2024large,pryzant2023automatic,yuksekgonul2025optimizing} score a candidate prompt using a single overall metric, so each iteration rewrites the entire prompt and can alter rules that were already working well. Because $\mathcal{M}_{\text{eval}}$ outputs the cited clause set $R(x_i; \mathcal{B})$ on both the baseline error set $\mathcal{E}_{0}(\mathcal{D}_{\text{train}})$ and the natural preservation set $\mathcal{P}_{0}(\mathcal{D}_{\text{train}})$, \textsc{LEDGER} assigns credit to each clause individually. For each clause $c_k \in \mathcal{B}$ with citation set $\mathcal{C}(c_k) = \{i \in \mathcal{D}_{\text{train}} : c_k \in R(x_i; \mathcal{B})\}$, intersecting $\mathcal{C}(c_k)$ with Equations~\eqref{eq:fix}--\eqref{eq:miss} gives the four-cell \emph{Clause Audit Ledger} $\mathcal{L}(c_k) = (\text{Fix}(c_k), \text{Preserve}(c_k), \text{Regress}(c_k), \text{Miss}(c_k))$, along with overall clause accuracy $\text{ClauseAcc}(c_k)$, transition fix precision $\text{FixPrec}(c_k)$, net accuracy gain $\text{NetGain}(c_k)$, and subgroup false positives fixed ($\text{SG-FP}_{\text{fixed}}(c_k)$) versus caused ($\text{SG-FP}_{\text{caused}}(c_k)$):
\begin{equation}
    \text{ClauseAcc}(c_k) = \frac{|\text{Fix}(c_k)| + |\text{Preserve}(c_k)|}{|\mathcal{C}(c_k)|}, \qquad
    \text{FixPrec}(c_k) = \frac{|\text{Fix}(c_k)|}{|\text{Fix}(c_k)| + |\text{Regress}(c_k)|},
    \label{eq:clause_acc}
\end{equation}
with $\text{NetGain}(c_k) = |\text{Fix}(c_k)| - |\text{Regress}(c_k)|$ as defined in Equation~\eqref{eq:clause_decomp}.

Using $\mathcal{L}(c_k)$, each subsequent optimization round updates clauses in $\mathcal{B}$ through three operations:
\begin{enumerate}
    \item \textbf{\textsc{Lock} (Freezing Verified Clauses):} A clause $c_k$ is assigned status \textsc{Lock} and kept unchanged in later rounds when it satisfies $\text{ClauseAcc}(c_k) \ge \tau_{\text{lock}}$ and $\text{FixPrec}(c_k) \ge \tau_{\text{lock}}$ (with $\tau_{\text{lock}} = 0.80$), achieves a positive net gain ($\text{NetGain}(c_k) > 0$), and causes no net increase in subgroup false positives ($\text{SG-FP}_{\text{caused}}(c_k) \le \text{SG-FP}_{\text{fixed}}(c_k)$). Requiring both $\text{ClauseAcc}(c_k) \ge \tau_{\text{lock}}$ and $\text{FixPrec}(c_k) \ge \tau_{\text{lock}}$ ensures that a clause is not locked merely because it agrees with the base policy on easy preservation examples while causing regressions on boundary cases. Because editing an upstream Tier~1 rule can shift which examples reach a downstream locked Tier~2 rule, the updated playbook is re-evaluated as a whole before acceptance.
    \item \textbf{\textsc{Prune} (Removing Inactive or Harmful Clauses):} \textsc{LEDGER} removes any unlocked clause that is never cited on $\mathcal{D}_{\text{train}}$ ($|\mathcal{C}(c_k)| = 0$) or causes a net accuracy loss without reducing subgroup false positives, while keeping at least one rule per active policy section and verdict so no category loses coverage.
    \item \textbf{\textsc{Repair} (Refining Exclusion Boundaries):} All remaining unlocked clauses are assigned status \textsc{Repair}. For each such clause $c_k$, \textsc{LEDGER} passes its fixed examples $\text{Fix}(c_k)$, its regressions $\text{Regress}(c_k)$, its persistent errors $\text{Miss}(c_k)$, and any uncited training errors to $\mathcal{M}_{\text{opt}}$. $\mathcal{M}_{\text{opt}}$ proposes refined versions of $c_k$ that tighten the exclusion boundary $\psi_k^{-}$ against $\text{Regress}(c_k)$ and adjust the trigger $\psi_k^{+}$ to cover missed errors while keeping all \textsc{Lock} clauses unchanged.
\end{enumerate}

\subsection{Candidate Selection Under Subgroup Fairness Constraints}
To select the best playbook across candidates and rounds, we convert the constrained objective in Equation~\eqref{eq:constrained_obj} into a penalized Lagrangian objective. For an evaluation split $\mathcal{D}$, let $\text{MinF1}(\mathcal{B}; \mathcal{D}) = \min_{y \in \mathcal{Y}} \text{F1}_y(\mathcal{B}; \mathcal{D})$ denote the worst-class F1 score. We score each candidate playbook $\mathcal{B}$ using:
\begin{align}
    U(\mathcal{B}; \mathcal{D}) &= \text{MacroF1}(\mathcal{B}; \mathcal{D}) - \lambda_{\text{FPR}} \max\bigl(0,\, \text{FPR}_{\text{SG}}(\mathcal{B}; \mathcal{D}) - \text{FPR}_{\text{SG}}(\emptyset; \mathcal{D}) - \delta_{\text{FPR}}\bigr) \nonumber \\
    &\quad - \lambda_{\text{deg}} \max\bigl(0,\, \text{MinF1}(\emptyset; \mathcal{D}) - \text{MinF1}(\mathcal{B}; \mathcal{D})\bigr),
    \label{eq:gov_score}
\end{align}
where the hinge terms penalize candidates that exceed the subgroup false-positive tolerance $\delta_{\text{FPR}}$ or degrade worst-class F1 below the base policy $P_0$ (with ties broken by higher $\text{MinF1}$ and lower $\text{FPR}_{\text{SG}}$). To reduce sampling variance when individual demographic subgroups have only a few examples on $\mathcal{D}_{\text{val}}$ alone, candidates are ranked by averaging $U(\mathcal{B}; \mathcal{D}_{\text{val}})$ and $U(\mathcal{B}; \mathcal{D}_{\text{train}})$, and a candidate from a later round is accepted only if it improves this score without violating the non-degradation constraints.

\section{Experimental Setup}
\label{sec:setup}

\subsection{Public Policy-Enforcement Benchmarks and Official Policy Specifications}
We evaluate \textsc{LEDGER} on three public benchmarks covering platform speech moderation, implied social-bias detection, and legal consumer contract compliance. For all three benchmarks, we construct the frozen Policy Prompt $P_0$ directly from the \emph{official policy taxonomy and category definitions} published by the benchmark authors, organized into three ordered sections: \textbf{Section~S1} ($s_1$, primary or substantive violation boundary), \textbf{Section~S2} ($s_2$, secondary or procedural violation boundary), and \textbf{Section~S3} ($s_3$, $y_{\text{default}}$, compliant outcome). Under Equation~\eqref{eq:precedence_order}, the playbook decision-list order is $\mathcal{T}_1^{\text{S1}} \succ \mathcal{T}_2^{\text{S1}} \succ \mathcal{T}_1^{\text{S2}} \succ \mathcal{T}_2^{\text{S2}} \succ P_0$. At test time, $\mathcal{M}_{\text{eval}}$ receives only the raw input text $x$ together with $(P_0, \mathcal{B})$. Using random seed $42$ for all stratified splits, we evaluate on:
\begin{enumerate}
    \item \textbf{HateXplain (Platform Speech Moderation)} \cite{mathew2021hatexplain}: Contains Twitter and Gab posts annotated into $\mathcal{Y} = \{\texttt{HATESPEECH}, \texttt{OFFENSIVE}, \texttt{NORMAL}\}$ under the benchmark's three-class speech policy (Section~S1 \texttt{HATESPEECH}: attacks or exclusion targeting a protected group; Section~S2 \texttt{OFFENSIVE}: profanity or insults that do not demean a protected group; Section~S3 \texttt{NORMAL}: benign speech), with target community tags and human rationales. We evaluate the eight most frequent demographic subgroups (\texttt{African}, \texttt{Islam}, \texttt{Homosexual}, \texttt{Jewish}, \texttt{Women}, \texttt{Refugee}, \texttt{Arab}, and \texttt{Caucasian}) using \emph{Subgroup Hate-FPR} (Equation~\ref{eq:fpr_sg}): the fraction of non-hateful posts mentioning a protected group that are falsely classified as \texttt{HATESPEECH}. Our stratified splits ($|\mathcal{D}_{\text{train}}| = 480$, $|\mathcal{D}_{\text{val}}| = 200$, $|\mathcal{D}_{\text{test}}| = 500$) match the class proportions of the $1{,}924$-post test corpus ($31.0\%$ \texttt{HATESPEECH}, $28.4\%$ \texttt{OFFENSIVE}, and $40.6\%$ \texttt{NORMAL}).
    \item \textbf{Social Bias Frames / SBIC.v2 (Implied Social-Bias Policy)} \cite{sap2020social}: Contains social media posts annotated for offensiveness (\texttt{offensiveYN}), group targeting (\texttt{whoTarget}), targeted minority group, and implied stereotypes (\texttt{targetStereotype}). Following \citet{sap2020social}, offensive posts targeting a demographic group map to Section~S1 (\texttt{HATESPEECH}), offensive posts without group targeting map to Section~S2 (\texttt{OFFENSIVE}), and non-offensive posts map to Section~S3 (\texttt{NORMAL}). Alongside three-class Macro-F1 and accuracy, we report positive-class F1 on the two binary SBIC subtasks (\emph{Is Offensive?} and \emph{Targeting Group?}) on $|\mathcal{D}_{\text{test}}| = 350$ across six demographic subgroups (\texttt{Black\_AAVE}, \texttt{Women\_Gender}, \texttt{Jewish}, \texttt{Muslim\_Islam}, \texttt{LGBTQ}, and \texttt{Disabled}), preserving the official test class distribution ($42.3\%$ \texttt{HATESPEECH}, $12.6\%$ \texttt{OFFENSIVE}, and $45.1\%$ \texttt{NORMAL}).
    \item \textbf{Unfair-ToS from CLAUDETTE / LexGLUE (Consumer Contract Compliance)} \cite{lippi2019claudette,chalkidis2022lexglue}: Contains consumer Terms of Service (ToS) clauses annotated by legal experts under EU Council Directive 93/13/EEC across eight statutory categories \cite{lippi2019claudette} mapped to $\mathcal{Y} = \{\texttt{SUBSTANTIVE\_UNFAIR}, \texttt{PROCEDURAL\_UNFAIR}, \texttt{COMPLIANT}\}$. \textbf{Section~S1} (\texttt{SUBSTANTIVE\_UNFAIR}) covers unilateral changes to obligations or liability: \emph{Limitation of liability} (\texttt{ltd}), \emph{Unilateral termination} (\texttt{ter}), \emph{Unilateral change} (\texttt{ch}), and \emph{Content removal} (\texttt{cr}). \textbf{Section~S2} (\texttt{PROCEDURAL\_UNFAIR}) covers dispute and consent restrictions: \emph{Mandatory arbitration} (\texttt{a}), \emph{Exclusive jurisdiction} (\texttt{j}), \emph{Choice of law} (\texttt{law}), and \emph{Contract by using / browsewrap} (\texttt{use}). \textbf{Section~S3} (\texttt{COMPLIANT}) covers fair clauses. Because unfair clauses make up $10.7\%$ of the full test corpus ($115$ substantive and $57$ procedural violations), our stratified splits ($|\mathcal{D}_{\text{train}}| = 480$, $|\mathcal{D}_{\text{val}}| = 200$, $|\mathcal{D}_{\text{test}}| = 350$) retain $99.4\%$ of all unfair test clauses ($115$ \texttt{SUBSTANTIVE\_UNFAIR} and $56$ \texttt{PROCEDURAL\_UNFAIR}) paired with $179$ \texttt{COMPLIANT} clauses containing contractual trigger terms, tracking \emph{Subgroup FPR} across six topic subgroups (\texttt{Liability\_Disclaimers}, \texttt{Termination\_Suspension}, \texttt{Modification\_Updates}, \texttt{Content\_Moderation}, \texttt{Dispute\_Forum\_Law}, and \texttt{Implied\_Consent}).
\end{enumerate}

\subsection{Models and Experimental Configurations}
To test within-pair performance and zero-shot cross-model transfer, we use two offline \emph{Optimizer Models} ($\mathcal{M}_{\text{opt}}$): \textbf{Gemini 4 Argon} (medium reasoning budget, temperature $0.4$) and \textbf{Claude Opus 5.5} (\texttt{claude-opus-5.5}, high reasoning effort); and three frozen single-pass \emph{Evaluation Models} ($\mathcal{M}_{\text{eval}}$) at temperature $0.0$: \textbf{Gemini 3.1 Flash-Lite} (\texttt{thinking\_mode=off}), \textbf{Gemma 4 26B-A4B} (\texttt{gemma-4-26b-a4b-it}, open-weights MoE with $26\text{B}$ total and $4\text{B}$ active parameters), and \textbf{Zhipu GLM-5.2} (\texttt{glm-5.2-fp8}). Playbooks are learned on $\mathcal{M}_{\text{eval}} = \text{Gemini 3.1 Flash-Lite}$ and evaluated zero-shot without re-optimization on Gemma~4 26B-A4B and Zhipu GLM-5.2.

In our main experiments, we compare (1) \textbf{Policy Prompt Only ($\emptyset$)}, where $\mathcal{M}_{\text{eval}}$ receives only the official three-section Policy Prompt $P_0$ and input $x$, against (2) \textbf{\textsc{LEDGER} (Ours)}, using fixed hyperparameters across all datasets and model pairs ($\tau_{\text{lock}} = 0.80$, $\delta_{\text{FPR}} = 0.10$; full details in Appendix~\ref{app:playbooks}.1) with non-parametric $95\%$ bootstrap confidence intervals ($10{,}000$ resamples). Appendix~\ref{app:ablations} provides ablations against flat error rules, whole-prompt rewrites, and Round~1 synthesis without Lock-Repair-Prune (Table~\ref{tab:ablation_main}), alongside contextual comparisons with published full-corpus supervised models (Table~\ref{tab:published_ref}).

\section{Empirical Results}
\label{sec:results}

\subsection{Performance on HateXplain and SBIC.v2}

\begin{table*}[t]
\centering
\caption{Held-out test performance on \textbf{HateXplain} ($N_{\text{test}}=500$, seed $42$) using Optimizer Model $\mathcal{M}_{\text{opt}} = \text{Gemini 4 Argon}$ and frozen Evaluation Model $\mathcal{M}_{\text{eval}} = \text{Gemini 3.1 Flash-Lite}$. Fixes ($W \to R$) and Collateral Regressions ($R \to W$) are exact-example transitions relative to \textbf{Policy Prompt Only}. Non-parametric $95\%$ bootstrap confidence intervals on Macro-F1: Policy Prompt Only $[54.34, 63.13]$, \textsc{LEDGER} $[64.06, 72.22]$.}
\label{tab:hatexplain_main}
\resizebox{\textwidth}{!}{%
\begin{tabular}{llcccccccc}
\toprule
\textbf{System / Method} & \textbf{Weight Updates / Split} & \textbf{Acc (\%)} & \textbf{Macro-F1 (\%)} & \textbf{Hate-F1} & \textbf{Off-F1} & \textbf{Norm-F1} & \textbf{Fixes / Regs} & \textbf{SG Hate-FPR (\%)} & \textbf{Cite (\%)} \\
\midrule
\textbf{Policy Prompt Only ($\emptyset$)} & \textbf{Zero ($0$ rules)} & 60.20 & 58.88 & 72.04 & 47.46 & 57.14 & $0 / 0$ & 40.88 & 0.0 \\
\textbf{\textsc{LEDGER} (Ours)} & \textbf{Zero ($10$ rules)} & \textbf{68.60} & \textbf{68.35} & \textbf{77.44} & \textbf{57.05} & \textbf{70.56} & $\mathbf{79} / \mathbf{37}$ & \textbf{16.79} & \textbf{99.6} \\
\bottomrule
\end{tabular}%
}
\end{table*}

Table~\ref{tab:hatexplain_main} reports held-out test performance on HateXplain ($N_{\text{test}}=500$). Using the frozen Policy Prompt alone, Gemini~3.1 Flash-Lite achieves $58.88\%$ Macro-F1 and $60.20\%$ accuracy, with a $40.88\%$ Subgroup Hate-FPR because it often flags reclaimed identity terms, AAE vernacular, and general insults as hate speech.

By organizing rules into two tiers within each policy section and enforcing ``do not apply when'' exclusion boundaries, \textsc{LEDGER} reaches $68.35\%$ test Macro-F1 ($+9.47$ points over Policy Prompt Only) and $68.60\%$ accuracy, fixing $79$ baseline errors against $37$ regressions ($87.71\%$ preservation retention). \textsc{LEDGER} improves F1 across all three classes ($+5.40$ points on \texttt{HATESPEECH}, $+9.59$ points on \texttt{OFFENSIVE}, and $+13.42$ points on \texttt{NORMAL}) while reducing Subgroup Hate-FPR by $24.09$ percentage points ($40.88\%$ to $16.79\%$) and citing a governing clause on $99.6\%$ of test predictions. Without updating model weights and using only $480$ training examples, \textsc{LEDGER} reaches the same Macro-F1 range as full-corpus fine-tuned BERT models ($67.40\%$--$68.70\%$ \cite{mathew2021hatexplain}; see Appendix~\ref{app:ablations}, Table~\ref{tab:published_ref}).

\begin{table*}[t]
\centering
\caption{Held-out test performance on \textbf{SBIC.v2} ($N_{\text{test}}=350$, seed $42$) across three-class policy enforcement (\texttt{HATESPEECH}, \texttt{OFFENSIVE}, \texttt{NORMAL}) and the standard binary SBIC subtasks (\emph{Is Offensive?} and \emph{Targeting Group?}). Non-parametric $95\%$ bootstrap confidence intervals on 3-Class Macro-F1: Policy Prompt Only $[64.40, 74.45]$, \textsc{LEDGER} $[68.62, 78.41]$.}
\label{tab:sbic_main}
\resizebox{\textwidth}{!}{%
\begin{tabular}{llccccccc}
\toprule
\textbf{System / Method} & \textbf{Paradigm / Rules} & \textbf{3-Class Acc} & \textbf{3-Class Macro-F1} & \textbf{Is Offensive? F1} & \textbf{Targeting Group? F1} & \textbf{Fixes / Regs} & \textbf{SG Hate-FPR} & \textbf{Cite (\%)} \\
\midrule
\textbf{Policy Prompt Only ($\emptyset$)} & \textbf{Zero ($0$ rules)} & 73.14 & 69.49 & 83.92 & 78.60 & $0 / 0$ & \textbf{7.41} & 0.0 \\
\textbf{\textsc{LEDGER} (Ours)} & \textbf{Zero ($10$ rules)} & \textbf{80.00} & \textbf{73.65} & \textbf{85.48} & \textbf{83.92} & $\mathbf{49} / \mathbf{25}$ & 18.52 & \textbf{96.6} \\
\bottomrule
\end{tabular}%
}
\end{table*}

Table~\ref{tab:sbic_main} presents held-out test results on SBIC.v2 ($N_{\text{test}}=350$). Unlike HateXplain, where Policy Prompt Only over-enforces \texttt{HATESPEECH}, baseline errors on SBIC.v2 are dominated $9\text{:}1$ by missed implied group bias ($E_{\text{under}}^{\text{S1}} \gg E_{\text{over}}^{\text{S1}}$, with a baseline Subgroup Hate-FPR of $7.41\%$ across the $27$ non-hateful test posts mentioning tracked subgroups). \textsc{LEDGER} allocates more capacity to Section~S1 Tier~2 violation rules to capture coded and implied stereotypes while maintaining a Tier~1 safe harbor (\texttt{S1.B2->NORMAL}) for benign peer vernacular. This raises three-class Macro-F1 from $69.49\%$ to $73.65\%$ ($+4.16$ points) and accuracy from $73.14\%$ to $80.00\%$ ($49$ fixes versus $25$ regressions), while improving binary \emph{Is Offensive?} F1 to $85.48\%$ and \emph{Targeting Group?} F1 from $78.60\%$ to $83.92\%$ ($+5.32$ points) with $96.6\%$ explicit clause citation (compared to $78.80\%$ and $69.90\%$ for fine-tuned BERT in \citet{sap2020social}; see Appendix~\ref{app:ablations}).

\subsection{Generalization to Legal Contract Compliance (Unfair-ToS / LexGLUE)}

\begin{table*}[t]
\centering
\caption{Held-out test performance on \textbf{Unfair-ToS (CLAUDETTE / LexGLUE)} ($N_{\text{test}}=350$, seed $42$, enforcing the official eight-category EU Council Directive 93/13/EEC consumer contract policy) using $\mathcal{M}_{\text{opt}} = \text{Gemini 4 Argon}$ and $\mathcal{M}_{\text{eval}} = \text{Gemini 3.1 Flash-Lite}$. Non-parametric $95\%$ bootstrap CIs on 3-Section Macro-F1: Policy Prompt Only $[70.09, 79.42]$, \textsc{LEDGER} $[82.07, 89.65]$.}
\label{tab:unfair_tos_main}
\resizebox{\textwidth}{!}{%
\begin{tabular}{llccccccccc}
\toprule
\textbf{System / Method} & \textbf{Paradigm / Rules} & \textbf{3-Sec Acc} & \textbf{3-Sec Macro-F1} & \textbf{Subst.\ F1} & \textbf{Proc.\ F1} & \textbf{Compl.\ F1} & \textbf{Binary Unfair F1} & \textbf{Fixes / Regs} & \textbf{Subgroup FPR (\%)} & \textbf{Cite (\%)} \\
\midrule
\textbf{Policy Prompt Only ($\emptyset$)} & \textbf{Zero ($0$ rules)} & 75.14 & 74.89 & 80.15 & 72.86 & 71.67 & 79.61 & $0 / 0$ & 19.15 & 0.0 \\
\textbf{\textsc{LEDGER} (Ours)} & \textbf{Zero ($11$ rules)} & \textbf{85.71} & \textbf{86.01} & \textbf{85.00} & \textbf{87.39} & \textbf{85.63} & \textbf{86.35} & $\mathbf{46} / \mathbf{9}$ & \textbf{9.79} & \textbf{88.0} \\
\bottomrule
\end{tabular}%
}
\end{table*}

To evaluate whether \textsc{LEDGER} generalizes beyond social media moderation without changing hyperparameters, Table~\ref{tab:unfair_tos_main} reports results on legal contract compliance (\textbf{Unfair-ToS}), comparing \textbf{Policy Prompt Only} and \textbf{\textsc{LEDGER}} on the identical $N_{\text{test}}=350$ split.

Under the frozen EU Directive 93/13/EEC Policy Prompt alone, Gemini~3.1 Flash-Lite achieves $74.89\%$ three-section Macro-F1 and $79.61\%$ binary unfairness F1, but suffers from high false-positive rates on lawful contract clauses ($94.74\%$ violation recall versus $68.64\%$ violation precision, $19.15\%$ Subgroup FPR). In particular, the base Policy Prompt misclassifies $51.1\%$ of lawful liability and scope-of-service clauses, $29.3\%$ of lawful contract-update notices, $21.6\%$ of user-conduct moderation provisions, and $17.6\%$ of bilateral termination clauses as unlawful under Section~S1, and often confuses browsewrap modification clauses with Section~S1 substantive changes rather than Section~S2 procedural consent.

Learning an 11-clause two-tier Decision Playbook with \textsc{LEDGER} increases held-out three-section Macro-F1 by $11.12$ points to $86.01\%$ and binary unfairness F1 to $86.35\%$, fixing $46$ baseline errors against only $9$ regressions ($96.58\%$ preservation retention). At the same time, \textsc{LEDGER} cuts Subgroup FPR nearly in half ($19.15\%$ to $9.79\%$), reducing false positives to $0.0\%$ on lawful \texttt{Termination\_Suspension} clauses, $4.9\%$ on \texttt{Modification\_Updates}, and $3.7\%$ on \texttt{Implied\_Consent}. The Evaluation Model cites a specific playbook clause on $88.0\%$ of test predictions, while the remaining $12.0\%$ are default \texttt{COMPLIANT} clauses that fall through to $P_0$.

\subsection{Cross-Model Generalization Across Optimizers and Evaluators}

\begin{table*}[t]
\centering
\caption{Cross-model generalization on \textbf{HateXplain} ($N_{\text{test}}=500$, seed $42$) across $2$ offline \textbf{Optimizer Models} ($\mathcal{M}_{\text{opt}}$) and $3$ frozen \textbf{Evaluation Models} ($\mathcal{M}_{\text{eval}}$). Playbooks learned on Gemini~3.1 Flash-Lite are transferred \textbf{zero-shot without re-optimization} to open-weights Gemma~4 26B-A4B and Zhipu GLM-5.2.}
\label{tab:cross_model}
\resizebox{\textwidth}{!}{%
\begin{tabular}{llccccccc}
\toprule
\textbf{Optimizer ($\mathcal{M}_{\text{opt}}$)} & \textbf{Evaluator ($\mathcal{M}_{\text{eval}}$)} & \textbf{Policy Prompt Only F1} & \textbf{\textsc{LEDGER} F1} & \textbf{\textsc{LEDGER} Lift} & \textbf{Fixes / Regs} & \textbf{SG Hate-FPR ($\emptyset \to \mathcal{B}$)} & \textbf{Cite (\%)} \\
\midrule
Gemini 4 Argon & Gemini 3.1 Flash-Lite & 58.88 & \textbf{68.35} & \textbf{+9.47 (+16.1\%)} & $79 / 37$ & $40.88 \to \mathbf{16.79}$ & 99.6 \\
Gemini 4 Argon & Gemma 4 26B-A4B (0-Shot) & 58.34 & \textbf{64.97} & \textbf{+6.63 (+11.4\%)} & $62 / 31$ & $48.91 \to 33.58$ & 97.8 \\
Gemini 4 Argon & Zhipu GLM-5.2 (0-Shot) & 52.89 & \textbf{67.85} & \textbf{+14.96 (+28.3\%)} & $95 / 30$ & $54.74 \to \mathbf{26.28}$ & 97.2 \\
\midrule
Claude Opus 5.5 & Gemini 3.1 Flash-Lite & 58.88 & \textbf{66.49} & \textbf{+7.61 (+12.9\%)} & $73 / 39$ & $40.88 \to 26.28$ & 99.0 \\
Claude Opus 5.5 & Gemma 4 26B-A4B (0-Shot) & 58.34 & \textbf{64.05} & \textbf{+5.71 (+9.8\%)} & $68 / 44$ & $48.91 \to 28.47$ & 97.4 \\
Claude Opus 5.5 & Zhipu GLM-5.2 (0-Shot) & 52.89 & \textbf{64.85} & \textbf{+11.96 (+22.6\%)} & $83 / 28$ & $54.74 \to 32.85$ & 95.2 \\
\bottomrule
\end{tabular}%
}
\end{table*}

\begin{figure*}[t]
\centering
\begin{tikzpicture}
\begin{axis}[
    ybar,
    bar width=11pt,
    width=0.92\textwidth,
    height=4.4cm,
    ylabel={Test Macro-F1 (\%)},
    symbolic x coords={Argon-FlashLite, Argon-Gemma4, Argon-GLM52, Opus-FlashLite, Opus-Gemma4, Opus-GLM52},
    xticklabels={
        {\shortstack{Argon $\to$\\Flash-Lite}},
        {\shortstack{Argon $\to$\\Gemma 4\\(0-Shot)}},
        {\shortstack{Argon $\to$\\GLM-5.2\\(0-Shot)}},
        {\shortstack{Opus 5.5 $\to$\\Flash-Lite}},
        {\shortstack{Opus 5.5 $\to$\\Gemma 4\\(0-Shot)}},
        {\shortstack{Opus 5.5 $\to$\\GLM-5.2\\(0-Shot)}}
    },
    xtick=data,
    xticklabel style={font=\footnotesize},
    ymin=48,
    ymax=73,
    ymajorgrids=true,
    grid style={dashed, gray!40},
    legend style={at={(0.5,1.03)}, anchor=south, legend columns=2, font=\footnotesize, draw=black!50},
    enlarge x limits=0.11
]
\addplot[fill=black!55, draw=black] coordinates {
    (Argon-FlashLite, 58.88) (Argon-Gemma4, 58.34) (Argon-GLM52, 52.89)
    (Opus-FlashLite, 58.88) (Opus-Gemma4, 58.34) (Opus-GLM52, 52.89)
};
\addplot[fill=red!75, draw=red!85!black] coordinates {
    (Argon-FlashLite, 68.35) (Argon-Gemma4, 64.97) (Argon-GLM52, 67.85)
    (Opus-FlashLite, 66.49) (Opus-Gemma4, 64.05) (Opus-GLM52, 64.85)
};
\legend{Policy Prompt Only ($\emptyset$), \textsc{LEDGER} (Ours)}
\end{axis}
\end{tikzpicture}
\caption{Held-out test Macro-F1 (\%) on HateXplain ($N_{\text{test}}=500$) across the $2 \times 3$ grid of offline Optimizer Models (Gemini~4 Argon and Claude Opus~5.5) and frozen Evaluation Models (Gemini~3.1 Flash-Lite, Gemma~4 26B-A4B, and Zhipu GLM-5.2). Across all six $(\mathcal{M}_{\text{opt}}, \mathcal{M}_{\text{eval}})$ pairs, \textsc{LEDGER} improves Macro-F1 by $+5.71$ to $+14.96$ points over Policy Prompt Only.}
\label{fig:cross_model}
\end{figure*}
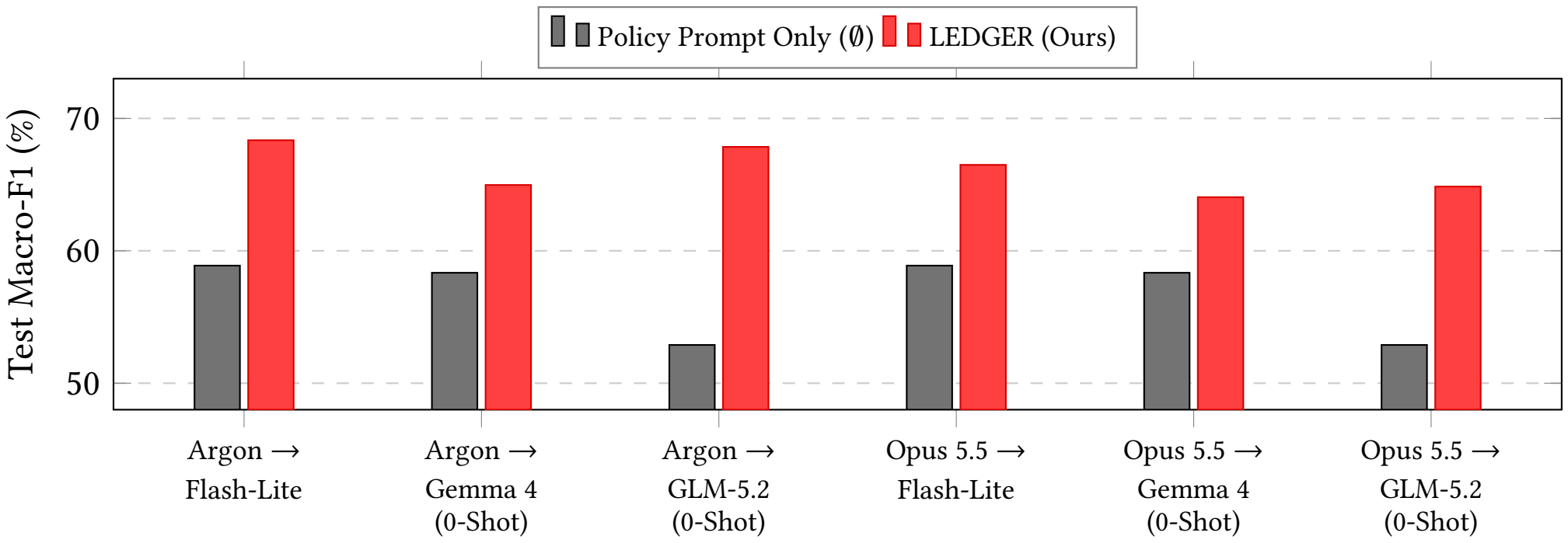

To test whether a learned Decision Playbook overfits to a single Optimizer Model or Evaluation Model, Table~\ref{tab:cross_model} and Figure~\ref{fig:cross_model} report results across all six $(\mathcal{M}_{\text{opt}}, \mathcal{M}_{\text{eval}})$ pairs on HateXplain ($N_{\text{test}}=500$). \textsc{LEDGER} improves performance over Policy Prompt Only across every $(\mathcal{M}_{\text{opt}}, \mathcal{M}_{\text{eval}})$ pair: when optimized by Gemini~4 Argon, \textsc{LEDGER} increases Macro-F1 by $+9.47$ points on Gemini~3.1 Flash-Lite ($58.88\%$ to $68.35\%$), $+6.63$ points zero-shot on open-weights Gemma~4 26B-A4B ($58.34\%$ to $64.97\%$), and $+14.96$ points zero-shot on Zhipu GLM-5.2 ($52.89\%$ to $67.85\%$); when optimized by Claude Opus~5.5, \textsc{LEDGER} improves Macro-F1 by $+7.61$ points on Gemini~3.1 Flash-Lite ($66.49\%$) and transfers zero-shot to Gemma~4 26B-A4B ($+5.71$ points, $64.05\%$) and Zhipu GLM-5.2 ($+11.96$ points, $64.85\%$). Across all six pairs, \textsc{LEDGER} fixes $62$--$95$ baseline test errors against $28$--$44$ regressions ($85.48\%$--$89.96\%$ preservation retention). Although Gemma~4 26B-A4B and Zhipu GLM-5.2 were never queried during optimization, both follow the JSON schema and cite clause IDs on $95.2\%$--$97.8\%$ of test posts while reducing Subgroup Hate-FPR by $14.60$--$28.46$ percentage points and Subgroup Benign-FPR by $10.00$--$15.00$ percentage points.

\subsection{Demographic Subgroup Fairness Analysis}

\begin{table*}[t]
\centering
\caption{Per-subgroup hate-speech false positive rate (\textbf{Subgroup Hate-FPR}, \%) on the held-out \textbf{HateXplain} test split ($N_{\text{test}}=500$) across all eight tracked protected demographic groups (lower is fairer; $n$ denotes the number of non-hateful test posts referencing each group). \textbf{Overall ($n=137$)} is the union of non-hateful test posts mentioning at least one tracked group (individual group denominators sum to $141$ because $4$ posts mention two groups). Subgroups with small test denominators (\texttt{Arab} $n=4$, \texttt{Caucasian} $n=9$, \texttt{Jewish} $n=11$) have wider per-post discretization steps ($25.0\%$, $11.1\%$, and $9.1\%$ per post) and should be interpreted alongside the aggregate rate ($n=137$).}
\label{tab:subgroup_fpr}
\resizebox{\textwidth}{!}{%
\begin{tabular}{lccccccccc}
\toprule
\textbf{Method ($\mathcal{M}_{\text{eval}}$)} & \textbf{Overall} & \textbf{African} & \textbf{Islam} & \textbf{Homosexual} & \textbf{Jewish} & \textbf{Women} & \textbf{Refugee} & \textbf{Arab} & \textbf{Caucasian} \\
 & ($n=137$) & ($n=23$) & ($n=19$) & ($n=29$) & ($n=11$) & ($n=29$) & ($n=17$) & ($n=4$) & ($n=9$) \\
\midrule
Policy Prompt Only (Flash-Lite) & 40.88 & 60.9 & 57.9 & 31.0 & 81.8 & 13.8 & 23.5 & 25.0 & 44.4 \\
\textbf{\textsc{LEDGER} (Argon $\to$ Flash-Lite)} & \textbf{16.79} & \textbf{34.8} & \textbf{21.1} & \textbf{0.0} & \textbf{63.6} & \textbf{0.0} & \textbf{0.0} & \textbf{0.0} & 22.2 \\
\textbf{\textsc{LEDGER} (Opus 5.5 $\to$ Flash-Lite)} & 26.28 & 60.9 & \textbf{21.1} & 20.7 & \textbf{63.6} & 6.9 & 5.9 & 25.0 & \textbf{11.1} \\
\midrule
Policy Prompt Only (GLM-5.2) & 54.74 & 69.6 & 79.0 & 48.3 & 72.7 & 27.6 & 47.1 & 50.0 & 66.7 \\
\textbf{\textsc{LEDGER} (Argon $\to$ GLM-5.2, 0-Shot)} & \textbf{26.28} & \textbf{43.5} & \textbf{42.1} & \textbf{6.9} & 63.6 & \textbf{13.8} & 11.8 & \textbf{25.0} & 55.6 \\
\textbf{\textsc{LEDGER} (Opus 5.5 $\to$ GLM-5.2, 0-Shot)} & 32.85 & 52.2 & 63.2 & 20.7 & \textbf{54.5} & \textbf{13.8} & \textbf{5.9} & 50.0 & \textbf{44.4} \\
\bottomrule
\end{tabular}%
}
\end{table*}

Table~\ref{tab:subgroup_fpr} breaks down Subgroup Hate-FPR across the eight tracked demographic groups on HateXplain. Under Policy Prompt Only, frozen classifiers frequently misclassify non-hateful posts mentioning minority groups: Gemini~3.1 Flash-Lite falsely labels $60.9\%$ of non-hateful \texttt{African} posts, $57.9\%$ of \texttt{Islam} posts, $81.8\%$ of \texttt{Jewish} posts, and $31.0\%$ of \texttt{Homosexual} posts as \texttt{HATESPEECH}, and Zhipu GLM-5.2 shows similarly high baseline rates ($69.6\%$ on \texttt{African}, $79.0\%$ on \texttt{Islam}, $48.3\%$ on \texttt{Homosexual}, $47.1\%$ on \texttt{Refugee}). Adding the Argon-optimized \textsc{LEDGER} playbook lowers overall Subgroup Hate-FPR on Gemini~3.1 Flash-Lite from $40.88\%$ ($56/137$) to $16.79\%$ ($23/137$): false positives drop from $60.9\%$ to $34.8\%$ on \texttt{African}, $57.9\%$ to $21.1\%$ on \texttt{Islam}, $31.0\%$ to $0.0\%$ on \texttt{Homosexual}, $13.8\%$ to $0.0\%$ on \texttt{Women}, and $23.5\%$ to $0.0\%$ on \texttt{Refugee}, while reducing false positives by $1$--$2$ posts on \texttt{Arab} ($n=4$) and \texttt{Caucasian} ($n=9$) and raising Hate-F1 from $72.04\%$ to $77.44\%$. Transferred zero-shot to Zhipu GLM-5.2, the same playbook reduces overall Subgroup Hate-FPR from $54.74\%$ to $26.28\%$ while raising GLM-5.2 Hate-F1 from $68.95\%$ to $79.55\%$.

\subsection{Clause-Level Audit Ledger and Lock-Repair-Prune Analysis}

\begin{table*}[t]
\centering
\caption{Four-cell \textbf{Clause Audit Ledger} $\mathcal{L}(c_k)$ on the HateXplain training split ($|\mathcal{D}_{\text{train}}|=480$) for the selected \textsc{LEDGER} Playbook ($\mathcal{M}_{\text{opt}}=\text{Gemini 4 Argon}, \mathcal{M}_{\text{eval}}=\text{Gemini 3.1 Flash-Lite}$). Each row records the number of cited training predictions in $\text{Right} = (\text{Fix} / \text{Preserve})$ and $\text{Wrong} = (\text{Regress} / \text{Miss})$, overall clause accuracy $\text{ClauseAcc}(c_k)$, transition fix precision $\text{FixPrec}(c_k) = |\text{Fix}|/(|\text{Fix}|+|\text{Regress}|)$, net accuracy gain, subgroup hate false positives fixed versus caused, and the audit status assigned by the locking rules in Section~\ref{sec:method}.3.}
\label{tab:clause_audit}
\resizebox{\textwidth}{!}{%
\begin{tabular}{llcccccccc}
\toprule
\textbf{Clause ID} & \textbf{Precedence Tier} & \textbf{Target Verdict} & \textbf{Right ($\text{Fix}/\text{Pres}$)} & \textbf{Wrong ($\text{Reg}/\text{Miss}$)} & \textbf{ClauseAcc (\%)} & \textbf{FixPrec (\%)} & \textbf{Net Gain} & \textbf{SG FP Fix / Cause} & \textbf{Audit Status} \\
\midrule
\texttt{S1.B1} & Tier 1 Safe-Harbor & \texttt{NORMAL} & $39\;(15 / 24)$ & $15\;(10 / 5)$ & 72.2 & 60.0 & $+5$ & $0 / 0$ & \textbf{\textsc{Repair}} \\
\texttt{S1.B2} & Tier 1 Safe-Harbor & \texttt{NORMAL} & $48\;(9 / 39)$ & $16\;(4 / 12)$ & 75.0 & 69.2 & $+5$ & $0 / 0$ & \textbf{\textsc{Repair}} \\
\texttt{S1.B3} & Tier 1 Safe-Harbor & \texttt{OFFENSIVE} & $38\;(10 / 28)$ & $20\;(6 / 14)$ & 65.5 & 62.5 & $+4$ & $8 / 0$ & \textbf{\textsc{Repair}} \\
\texttt{S1.B4} & Tier 1 Safe-Harbor & \texttt{OFFENSIVE} & $30\;(16 / 14)$ & $33\;(8 / 25)$ & 47.6 & 66.7 & $+8$ & $14 / 0$ & \textbf{\textsc{Repair}} \\
\texttt{S1.C1} & Tier 2 Inclusion & \texttt{HATESPEECH} & $96\;(3 / 93)$ & $21\;(2 / 19)$ & 82.1 & 60.0 & $+1$ & $0 / 2$ & \textbf{\textsc{Repair}} \\
\texttt{S1.C2} & Tier 2 Inclusion & \texttt{HATESPEECH} & $49\;(0 / 49)$ & $18\;(0 / 18)$ & 73.1 & 100.0 & $+0$ & $0 / 0$ & \textbf{\textsc{Repair}} \\
\midrule
\texttt{S2.B1} & Tier 1 Safe-Harbor & \texttt{NORMAL} & $42\;(19 / 23)$ & $5\;(3 / 2)$ & \textbf{89.4} & \textbf{86.4} & \textbf{$+16$} & $0 / 0$ & \textbf{\textsc{Lock}} \\
\texttt{S2.B2} & Tier 1 Safe-Harbor & \texttt{NORMAL} & $15\;(6 / 9)$ & $3\;(3 / 0)$ & 83.3 & 66.7 & $+3$ & $0 / 0$ & \textbf{\textsc{Repair}} \\
\texttt{S2.C1} & Tier 2 Inclusion & \texttt{OFFENSIVE} & $23\;(1 / 22)$ & $17\;(1 / 16)$ & 57.5 & 50.0 & $+0$ & $0 / 0$ & \textbf{\textsc{Repair}} \\
\texttt{S2.C2} & Tier 2 Inclusion & \texttt{OFFENSIVE} & $6\;(1 / 5)$ & $3\;(0 / 3)$ & 66.7 & 100.0 & $+1$ & $1 / 0$ & \textbf{\textsc{Repair}} \\
\bottomrule
\end{tabular}%
}
\end{table*}

Table~\ref{tab:clause_audit} shows the Clause Audit Ledger $\mathcal{L}(c_k)$ on HateXplain ($\mathcal{D}_{\text{train}}=480$, $543$ total clause citations, or $1.13$ per post). Every clause has $\text{NetGain}(c_k) \ge 0$, showing that two-tier precedence and exclusion boundaries prevent individual clauses from causing net regressions:
\begin{itemize}
    \item \textbf{Distinguishing \texttt{S2.B1} from \texttt{S1.C1} and \texttt{S2.B2}:} Locking requires both $\text{ClauseAcc}(c_k) \ge 80.0\%$ and $\text{FixPrec}(c_k) \ge 80.0\%$, alongside $\text{NetGain} > 0$ and no new subgroup false positives. Clause \texttt{S2.B1} satisfies all criteria ($\text{ClauseAcc} = 89.4\%$, $\text{FixPrec} = 86.4\%$, $\text{NetGain} = +16$, $\text{SG-FP}_{\text{caused}} = 0$) and is locked. Although \texttt{S1.C1} and \texttt{S2.B2} exceed $80\%$ $\text{ClauseAcc}$ ($82.1\%$ and $83.3\%$) on preserved examples, their fix precision on changed predictions is below $80\%$ (\texttt{S1.C1} has $\text{FixPrec} = 60.0\%$ and causes $2$ subgroup false positives; \texttt{S2.B2} has $\text{FixPrec} = 66.7\%$), so both are assigned \textsc{Repair}.
    \item \textbf{Correcting Subgroup False Positives and Multi-Round Gains:} In Section~S1 Tier~1, clauses \texttt{S1.B3} and \texttt{S1.B4} fix $22$ baseline Subgroup Hate-FPR errors ($8$ by \texttt{S1.B3} and $14$ by \texttt{S1.B4}) without causing any new subgroup false positives. While Round~1 synthesis reaches the validation optimum on HateXplain ($480/200$), Round~2 Lock-Repair-Prune improves test Macro-F1 by $+4.88$ points on HateXplain ($240/100$, locking $4$ clauses and cutting Subgroup Hate-FPR from $35.77\%$ to $21.17\%$), $+1.84$ points on SBIC.v2 ($480/200$, locking $3$ clauses and pruning $1$ inactive clause), and $+2.21$ points on Unfair-ToS ($480/200$, locking $6$ clauses and lowering Subgroup FPR from $12.34\%$ to $9.79\%$).
\end{itemize}

\section{Discussion, Governance Implications, and Limitations}
\label{sec:discussion}

\paragraph{Separating Official Policy from Operational Playbooks.}
In legal compliance, safety review, healthcare administration, and platform governance, official policy documents must remain stable and understandable to users, while day-to-day enforcement requires detailed rules for edge cases, exceptions, and domain-specific language \cite{gorwa2020algorithmic,vaccaro2020contesting}. By keeping the official Policy Prompt $P_0$ frozen and learning a separate Decision Playbook $\mathcal{B}$, \textsc{LEDGER} separates high-level policy definitions from operational guidance. Policy authors, legal counsel, and auditors can review each clause $c_k \in \mathcal{B}$ alongside its training ledger $\mathcal{L}(c_k)$ before deployment, verify how each Tier~1 exception rule affects protected subgroups, and lock legally required rules against automated edits.

\paragraph{Contestability and Self-Reported Rule Citations.}
Across all three benchmarks and all evaluated models, the Evaluation Model cites explicit clause identifiers (\texttt{applied\_rule\_ids}) on $88.0\%$ to $100.0\%$ of test predictions, linking each decision to a specific playbook clause. Because $\mathcal{M}_{\text{eval}}$ is a language model processing the playbook in a single forward pass rather than a symbolic interpreter, \texttt{applied\_rule\_ids} represents self-reported rule attribution rather than a formal proof of causal execution. Even so, our training and validation audits show that editing or pruning a clause $c_k$ changes predictions primarily on its citation set $\mathcal{C}(c_k)$, providing a practical audit trail when organizations revise or retire individual rules.

\paragraph{Limitations and Future Work.}
Our study has several limitations that suggest directions for future work:
\begin{itemize}
    \item \textbf{Training Subgroup Metadata and Annotator Notes:} Computing Equation~\eqref{eq:gov_score} uses subgroup indicators $G_i$ on $\mathcal{D}_{\text{train}}$ and $\mathcal{D}_{\text{val}}$ (from annotator tags or statutory/identity lexicons), and training error prompts include dataset rationales or subcategory notes when available (shared equally across all playbook baselines in Appendix~\ref{app:ablations}). When unavailable, subgroup tags can be approximated via lexicons or an auxiliary classifier, and future work should ablate learning playbooks from raw text and coarse labels alone.
    \item \textbf{Language Coverage:} All three benchmarks are English-language text datasets, and a $10$--$11$ clause playbook adds roughly $550$--$700$ static prefix tokens per request (amortizable via prefix caching). Extending \textsc{LEDGER} to multilingual and multimodal policies remains future work.
\end{itemize}

\section{Conclusion}
\label{sec:conclusion}

Enforcing written policies with frozen language models requires bridging the gap between general policy principles and recurring edge cases. When practitioners append flat lists of corrective rules to a Policy Prompt, new rules often conflict with existing ones and overturn examples the base policy already classified correctly. \textsc{LEDGER} addresses this problem by organizing rules within each policy section into a two-tier decision list that checks exception rules before violation rules, pairing each rule with an explicit boundary on when it does not apply. By requiring the model to cite the active rule for each prediction during training, \textsc{LEDGER} tracks each rule's fixes and regressions to lock reliable rules, revise rules that cause errors or subgroup false positives, and prune unused rules. Across hate speech moderation, implied social-bias detection, and consumer contract compliance, our results show that structured rule precedence and rule-level auditing allow frozen language models to improve accuracy and reduce subgroup false positives in a single pass while producing an auditable playbook that transfers across model families.

\section*{Ethical Considerations Statement}
This study evaluates automated policy enforcement using three publicly available, peer-reviewed benchmarks: HateXplain \cite{mathew2021hatexplain}, the Social Bias Inference Corpus (SBIC.v2) \cite{sap2020social}, and LexGLUE Unfair-ToS \cite{lippi2019claudette,chalkidis2022lexglue}. No new human-subjects data collection, crowdworker annotation, or scraping of private user data was conducted for this work. Because HateXplain and SBIC.v2 contain offensive language, slurs, and expressions of social bias, we avoid reproducing gratuitous slurs in the text of the paper and limit our verbatim playbook examples to abstract policy criteria and structural boundaries. A central ethical motivation of our work is mitigating the documented tendency of automated hate-speech classifiers to over-flag non-hateful speech by or about marginalized demographic groups, including reclaimed slurs, dialectal variation, counterspeech, and community discussion of discrimination \cite{sap2019risk,davidson2019racial,dixon2018measuring}. Finally, we emphasize that \textsc{LEDGER} is designed to support human-in-the-loop policy governance rather than replace human oversight: learned playbooks should be reviewed by domain experts and affected stakeholders before deployment.

\section*{Generative AI Usage Statement}
The authors used AI tools for editing and grammar checking and take full responsibility for the content of this paper.

\bibliographystyle{ACM-Reference-Format}
\bibliography{references}

\appendix

\section{Verbatim Policy Prompts, Output Schema, and Learned Decision Playbooks}
\label{app:playbooks}

\subsection{Implementation Details, Evaluation Prompt Structure, and JSON Output Schema}
\paragraph{Optimization Hyperparameters.}
Across all benchmarks and model pairs, \textsc{LEDGER} uses the same fixed settings without per-dataset tuning: $K_{\max} = 10$--$12$ total clauses (allocating $3$--$4$ clauses to $\mathcal{T}_1^{(s_1)}$ and $2$--$4$ to $\mathcal{T}_2^{(s_1)}$ when $E_{\text{over}}^{(s_1)} \ge E_{\text{under}}^{(s_1)}$, or $1$--$2$ to $\mathcal{T}_1^{(s_1)}$ and $3$--$4$ to $\mathcal{T}_2^{(s_1)}$ when $E_{\text{under}}^{(s_1)} > E_{\text{over}}^{(s_1)}$, plus $1$--$2$ clauses each in $\mathcal{T}_1^{(s_2)}$ and $\mathcal{T}_2^{(s_2)}$), $3$ initial candidates in Round~1, $120$ stratified preservation examples from $\mathcal{P}_{0}(\mathcal{D}_{\text{train}})$, and $2$ repair candidates per round across $T=3$ rounds on HateXplain/SBIC.v2 and $T=2$ rounds on Unfair-ToS. In the Clause Audit Ledger (Section~\ref{sec:method}.3), the locking threshold is $\tau_{\text{lock}} = 0.80$ (with a minimum support of $2$ fixed or preserved examples), and one highest-accuracy clause per active $(s_k, y_k)$ pair is retained so pruning never leaves a policy category uncovered. In the Lagrangian selection objective (Equation~\ref{eq:gov_score}), the constraint penalty weights are $\lambda_{\text{FPR}} = \lambda_{\text{deg}} = 0.25$ and the subgroup false-positive tolerance is $\delta_{\text{FPR}} = 0.10$ (or $\delta_{\text{FPR}} = 0.0$ in our strict-fairness ablation).

\paragraph{Evaluation Prompt Structure and JSON Output Schema.}
At evaluation time, the frozen Evaluation Model $\mathcal{M}_{\text{eval}}$ receives the Frozen Official Policy Prompt $P_0$, the two-tier Decision Playbook $\mathcal{B}$ (omitted under Policy Prompt Only), the input text $x$, and the following output instructions requiring explicit clause citation (\texttt{applied\_rule\_ids}, abbreviated \texttt{rule\_ids} in the main text):
{\footnotesize\begin{verbatim}
### OUTPUT INSTRUCTIONS
Classify the post above by following the Official Policy and the Detailed
  Policy Interpretation Playbook.
Return ONLY a single-line valid JSON object with keys:
  "rationale": brief 1-sentence explanation citing the decisive policy boundary,
  "applied_rule_ids": list of exact Playbook rule IDs (e.g. ["S1.B1"] or [])
    whose conditions directly determined your verdict,
  "verdict": "<SECTION_S1_VERDICT|SECTION_S2_VERDICT|SECTION_S3_VERDICT>"
\end{verbatim}}

\subsection{HateXplain Frozen Official Policy Prompt and Learned LEDGER Playbooks}

\paragraph{Frozen Official Policy Prompt ($P_0$).}
{\footnotesize\begin{verbatim}
You are a content moderation classifier enforcing an official
  three-category platform speech
policy. Every post must be assigned to exactly one policy category:
  HATESPEECH, OFFENSIVE, or
NORMAL.

Section S1 - Hate Speech (HATESPEECH):
Assign HATESPEECH to content that attacks, demeans, dehumanizes, degrades,
  or incites hatred,
exclusion, or violence against people based on a protected demographic
  characteristic (such as
race, ethnicity, nationality, religion, gender, sexual orientation,
  disability, or
immigrant/refugee status).

Section S2 - Offensive Speech (OFFENSIVE):
Assign OFFENSIVE to content containing profanity, vulgarity, insults, or
  abusive/harassing
language directed at individuals, groups, or situations without attacking
  or demeaning a
protected demographic group.

Section S3 - Normal Speech (NORMAL):
Assign NORMAL to benign, neutral, informational, or everyday
  conversational posts that do not
violate Section S1 or Section S2.
\end{verbatim}}

\paragraph{Learned Two-Tier Decision Playbook ($480/200$ Split, $\mathcal{M}_{\text{opt}} = \text{Gemini 4 Argon}$, $68.35\%$ Test Macro-F1, $16.79\%$ Subgroup Hate-FPR).}
{\footnotesize\begin{verbatim}
### DETAILED POLICY INTERPRETATION PLAYBOOK (STRICT TWO-TIER PRECEDENCE)
This Playbook provides the detailed operational interpretation of the
  Official Policy above.
Apply it sequentially:
1. Protected-Group Boundary (Section S1): Whenever a post mentions,
  addresses, or alludes to a
protected demographic group or identity dialect, evaluate Section S1
  FIRST. Check S1 Tier 1
Safe-Harbor Clauses FIRST; apply an S1 Tier 1 clause ONLY when its
  positive context holds AND
none of its 'do not apply when' disqualifiers are triggered; otherwise
  evaluate S1 Tier 2
Hate-Speech Inclusion Clauses.
2. Non-Demographic Offensive vs. Normal Boundary (Section S2): Whenever
  HATESPEECH does not
apply, evaluate Section S2 to distinguish OFFENSIVE from NORMAL. Apply S2
  Tier 1 (NORMAL) ONLY
when profanity, slang, or blunt commentary lacks targeted personal abuse
  or severe lewd/shock
degradation AND none of its 'do not apply when' disqualifiers apply; if
  any S2 Tier 2 targeted
insult, harassment, or severe lewd/shock offense applies, assign
  OFFENSIVE; otherwise follow
the Official Policy.

#### Section S1 - Protected-Group & Social-Bias Boundary
  Tier 1 - Safe-Harbor Overrides (Check FIRST; overrides Tier 2 in S1):
    - [S1.B1] -> NORMAL: Assign NORMAL when protected demographic terms,
      reclaimed vernacular,
      or identity labels appear in benign in-group camaraderie, casual
      observational commentary,
      self-reference, or counter-speech condemning bigotry; do not apply
      when the post expresses
      hostility, exclusion, or degradation toward a demographic group.
    - [S1.B2] -> NORMAL: Assign NORMAL when a post discusses geopolitical
      conflicts, immigration
      enforcement, historical events, or religious and social policies
      without using demographic
      slurs or demeaning generalizations; do not apply when the post uses
      sarcastic mockery or
      hostile stereotypes to disparage a demographic group.
    - [S1.B3] -> OFFENSIVE: Assign OFFENSIVE when homophobic, ableist,
      socioeconomic, or
      gendered epithets are used as casual interpersonal insults,
      dismissive banter, or vulgar
      expressions of annoyance without promoting systemic hatred or
      dehumanization of a
      protected class; do not apply when combined with explicit racial or
      ethnic slurs or calls
      for group exclusion.
    - [S1.B4] -> OFFENSIVE: Assign OFFENSIVE when a post expresses
      abrasive grievances, cultural
      or socioeconomic stereotyping, or partisan hostility regarding
      demographic or migrant
      groups without inciting violence, alleging subhuman status, or
      deploying severe racial
      slurs; do not apply when the content frames the demographic group as
      vermin, inherent
      criminals, or targets for eradication.
  Tier 2 - Diagnostic Inclusion Criteria (Check SECOND if no Tier 1
    override in S1 applies):
    - [S1.C1] -> HATESPEECH: Assign HATESPEECH when content attacks an
      individual or group using
      severe racial, ethnic, antisemitic, religious, or transphobic slurs
      coupled with malice,
      sexual degradation, or expressions of hatred; do not apply when
      reclaimed dialect is used
      non-maliciously in everyday peer anecdotes.
    - [S1.C2] -> HATESPEECH: Assign HATESPEECH when content dehumanizes a
      protected demographic
      group, compares them to animals or filth, minimizes historical
      atrocities against them, or
      incites violence, expulsion, or legal subjugation against them; do
      not apply to
      non-violent political criticism of state actors or ideological
      movements.

#### Section S2 - Non-Demographic Offensive vs. Normal Boundary
  Tier 1 - Safe-Harbor Overrides (Check FIRST; overrides Tier 2 in S2):
    - [S2.B1] -> NORMAL: Assign NORMAL when profanity, vulgar slang, or
      ableist terms are used
      for emotional emphasis, self-deprecation, playful affection,
      fictional or media
      commentary, or casual storytelling without attacking a specific
      person; do not apply when
      the language is weaponized to insult or demean a targeted
      individual.
    - [S2.B2] -> NORMAL: Assign NORMAL when content references sensitive
      topics such as sexual
      assault, violent crime, political extremism, or public scandals in a
      reporting,
      analytical, or conversational manner; do not apply when the speaker
      wishes harm upon a
      target or uses graphic sexual violence to harass someone.
  Tier 2 - Diagnostic Inclusion Criteria (Check SECOND if no Tier 1
    override in S2 applies):
    - [S2.C1] -> OFFENSIVE: Assign OFFENSIVE when content directs abusive
      personal insults,
      ableist name-calling, vulgar hostility, or wishes of violent harm at
      an individual, public
      figure, or political group; do not apply when the attack relies on
      severe racial or
      religious slurs that violate Section S1.
    - [S2.C2] -> OFFENSIVE: Assign OFFENSIVE when content employs crude
      sexual degradation,
      demeaning generalizations about gender roles, or mocking shock humor
      about sexual violence
      or political tragedies without attacking a protected group under
      Section S1; do not apply
      to benign everyday complaining or neutral factual discussion.
\end{verbatim}}

\subsection{SBIC.v2 Frozen Official Policy Prompt and Learned LEDGER Playbooks}

\paragraph{Frozen Official Policy Prompt ($P_0$, Social Bias Inference Corpus v2 Standard).}
{\footnotesize\begin{verbatim}
You are a content moderation classifier enforcing an official three-category
  social-bias and platform speech policy. Every post must be assigned to
  exactly one policy category: HATESPEECH, OFFENSIVE, or NORMAL.

Section S1 - Group-Targeted Bias & Hate Speech (HATESPEECH):
Assign HATESPEECH to content that attacks, demeans, dehumanizes, or promotes
  derogatory bias or stereotypes against a protected demographic or
  marginalized group (such as race, ethnicity, gender, religion, sexual
  orientation, or disability).

Section S2 - Non-Demographic Offensive Speech (OFFENSIVE):
Assign OFFENSIVE to content containing profanity, vulgarity, or personal
  insults and attacks directed at individuals, behaviors, or situations
  without targeting a protected group.

Section S3 - Normal Speech (NORMAL):
Assign NORMAL to benign, neutral, informational, or everyday conversational
  posts that do not violate Section S1 or Section S2.
\end{verbatim}}

\paragraph{Learned Two-Tier Decision Playbook (\textsc{LEDGER}-Balanced, $240/100$ Split, $73.65\%$ Test Macro-F1, $18.52\%$ Subgroup Hate-FPR).}
{\footnotesize\begin{verbatim}
### DETAILED POLICY INTERPRETATION PLAYBOOK (STRICT TWO-TIER PRECEDENCE)
This Playbook provides the detailed operational interpretation of the
  Official Policy above. Apply it sequentially:
1. Protected-Group Boundary (Section S1): Whenever a post mentions,
  addresses, or alludes to a protected demographic group or identity
  dialect, evaluate Section S1 FIRST. Check S1 Tier 1 Safe-Harbor Clauses
  FIRST; apply an S1 Tier 1 clause ONLY if all of its non-derogation
  conditions hold; otherwise evaluate S1 Tier 2 Hate-Speech Inclusion
  Clauses.
2. Non-Demographic Offensive vs. Normal Boundary (Section S2): Whenever
  HATESPEECH does not apply, evaluate Section S2 to distinguish OFFENSIVE
  from NORMAL. Apply S2 Tier 1 (NORMAL) ONLY when language is non-targeted
  and benign; if the post directs personal insults, harassment, or severe
  lewd/shock vulgarity covered by S2 Tier 2, assign OFFENSIVE; otherwise
  follow the Official Policy.

#### Section S1 - Protected-Group & Social-Bias Boundary
  Tier 1 - Safe-Harbor Overrides (Check FIRST; overrides Tier 2 in S1):
    - [S1.B1] -> NORMAL: Assign NORMAL when a post references demographic
      identities, subcultures, or historical events in a neutral, factual,
      self-identificatory, or mildly self-deprecating manner without
      promoting hostility or derogatory stereotypes; do not apply when the
      reference implies group inferiority, criminality, or dehumanization.
    - [S1.B2] -> NORMAL: Assign NORMAL when reclaimed dialect, casual
      vernacular, or gendered slang is used purely for self-reference,
      friendly peer banter, or sports/pop-culture hype without demeaning a
      demographic group; do not apply when the language is used to
      generalize about, objectify, or disparage women or racial minorities
      as a class.
    - [S1.B3] -> OFFENSIVE: Assign OFFENSIVE when a post uses extreme taboo
      shock humor, sexual vulgarity, or provocative extremist
      self-aggrandizement directed at specific individuals, personal
      fantasies, or non-demographic situations; do not apply when the
      content mocks, stereotypes, or dehumanizes a protected demographic
      group or class of mass-tragedy victims.
  Tier 2 - Diagnostic Inclusion Criteria (Check SECOND if no Tier 1
    override in S1 applies):
    - [S1.C1] -> HATESPEECH: Assign HATESPEECH when a post uses overt or
      colloquial slurs, objectifying labels, or incel terminology to
      generalize about the sexual promiscuity, loyalty, intelligence, or
      moral worth of women, racial minorities, LGBTQ individuals, or
      religious groups; do not apply when a vulgar term is aimed solely at a
      single individual's non-demographic behavior in a personal dispute.
    - [S1.C2] -> HATESPEECH: Assign HATESPEECH when a post uses riddles,
      puns, setup-punchline jokes, or sarcastic observations to invoke
      derogatory stereotypes associating a racial, ethnic, religious,
      regional, or socioeconomic group with criminality, animalistic traits,
      abuse, or inferiority; do not apply when wordplay is entirely benign
      and free of demographic association.
    - [S1.C3] -> HATESPEECH: Assign HATESPEECH when a post demeans, mocks,
      or stereotypes people based on physical or mental disability, body
      weight, feminist advocacy, or status as victims of sexual assault,
      domestic abuse, or mass-casualty atrocities; do not apply when a post
      merely reports factual news about a tragedy or criticizes a specific
      public figure's actions without group-level bias.

#### Section S2 - Non-Demographic Offensive vs. Normal Boundary
  Tier 1 - Safe-Harbor Overrides (Check FIRST; overrides Tier 2 in S2):
    - [S2.B1] -> NORMAL: Assign NORMAL when profanity or vulgar intensifiers
      are used to express personal frustration, excitement, physical
      discomfort, or commentary on entertainment, objects, or personal life
      without attacking another person; do not apply when the profanity is
      weaponized to insult, demean, or harass a specific target.
    - [S2.B2] -> NORMAL: Assign NORMAL when a post recounts third-party
      anecdotes, quotes song lyrics or fictional dialogue, engages in
      consensual roleplay banter, or asks general rhetorical questions about
      human flaws without maliciously attacking a real person; do not apply
      when the post delivers a genuine interpersonal insult or crude sexual
      degradation.
  Tier 2 - Diagnostic Inclusion Criteria (Check SECOND if no Tier 1
    override in S2 applies):
    - [S2.C1] -> OFFENSIVE: Assign OFFENSIVE when a post directs hostile
      name-calling, aggressive personal insults, or harsh
      partisan/professional attacks at an individual, public figure, or
      organization without invoking protected-group bias; do not apply when
      the remark is mild, non-profane criticism of a product, sports team,
      or policy.
    - [S2.C2] -> OFFENSIVE: Assign OFFENSIVE when a post features crude
      sexual vulgarity, morbid puns about a specific individual's accidental
      death or personal tragedy, or grotesque shock humor that does not
      stereotype a protected demographic group; do not apply when the humor
      targets a demographic class or protected group.
\end{verbatim}}

\paragraph{Learned Two-Tier Decision Playbook (\textsc{LEDGER}-Flow Round 2, $480/200$ Split, $72.49\%$ Test Macro-F1, $85.81\%$ Group-Targeting F1).}
{\footnotesize\begin{verbatim}
### DETAILED POLICY INTERPRETATION PLAYBOOK (STRICT TWO-TIER PRECEDENCE)
This Playbook provides the detailed operational interpretation of the
  Official Policy above. Apply it sequentially:
1. Protected-Group Boundary (Section S1): Whenever a post mentions,
  addresses, or alludes to a protected demographic group or identity
  dialect, evaluate Section S1 FIRST. Check S1 Tier 1 Safe-Harbor Clauses
  FIRST; apply an S1 Tier 1 clause ONLY when its positive context holds AND
  none of its 'do not apply when' disqualifiers are triggered; otherwise
  evaluate S1 Tier 2 Hate-Speech Inclusion Clauses.
2. Non-Demographic Offensive vs. Normal Boundary (Section S2): Whenever
  HATESPEECH does not apply, evaluate Section S2 to distinguish OFFENSIVE
  from NORMAL. Apply S2 Tier 1 (NORMAL) ONLY when profanity, slang, or blunt
  commentary lacks targeted personal abuse or severe lewd/shock degradation
  AND none of its 'do not apply when' disqualifiers apply; if any S2 Tier 2
  targeted insult, harassment, or severe lewd/shock offense applies, assign
  OFFENSIVE; otherwise follow the Official Policy.

#### Section S1 - Protected-Group & Social-Bias Boundary
  Tier 1 - Safe-Harbor Overrides (Check FIRST; overrides Tier 2 in S1):
    - [S1.B1] -> NORMAL: Assign NORMAL when protected demographic
      identities or reclaimed
      in-group terms appear in purely benign self-identification, neutral
      informational updates,
      balanced policy reflection, or harmless everyday anecdotes; do not
      apply when the post
      uses phonetic slurs, expresses past or present dislike toward a
      group, argues against
      minority representation, or generalizes about a demographic.
    - [S1.B2] -> OFFENSIVE: Assign OFFENSIVE when explicit profanity,
      morbid death wishes, or
      dehumanizing personal insults are directed at a single named
      political figure or public
      official; do not apply to non-vulgar political commentary and
      partisan accusations, or to
      collective slurs and jokes stereotyping ideological groups or
      domestic violence victims.
  Tier 2 - Diagnostic Inclusion Criteria (Check SECOND if no Tier 1
    override in S1 applies):
    - [S1.C1] -> HATESPEECH: Assign HATESPEECH when third-person
      commentary, lyrics, or
      narrative jokes use gendered or racial slurs to objectify women or
      minorities as
      promiscuous, disloyal, or interchangeable, or attribute criminality,
      abuse, predation, or
      inferiority to a racial, national, religious, or sexual-orientation
      group; do not apply to
      direct second-person insults lacking group generalization.
    - [S1.C2] -> HATESPEECH: Assign HATESPEECH when setup-punchline
      riddles, puns, wordplay, or
      rhetorical questions encode implied derogatory stereotypes,
      historical atrocities, or
      subhuman associations against a protected demographic, nationality,
      or regional
      population, even without profanity; do not apply to benign wordplay
      lacking group bias.
    - [S1.C3] -> HATESPEECH: Assign HATESPEECH when content mocks,
      demeans, or promotes
      derogatory stereotypes targeting physical or mental disability,
      serious illness, body
      weight, gender-identity or feminist advocacy, or lower socioeconomic
      status; do not apply
      when disability or health terms are used in neutral or supportive
      contexts.
    - [S1.C4] -> HATESPEECH: Assign HATESPEECH when dark humor, analogies,
      or threats trivialize
      or encourage suicide, domestic or gender-based violence against
      women, sexual assault
      victim-blaming, or mass-casualty terrorist and genocidal atrocities;
      do not apply to
      first-person lewd taboo shock jokes or generic disaster and
      dead-child riddles lacking
      demographic or gender-violence bias.

#### Section S2 - Non-Demographic Offensive vs. Normal Boundary
  Tier 1 - Safe-Harbor Overrides (Check FIRST; overrides Tier 2 in S2):
    - [S2.B1] -> NORMAL: Assign NORMAL when non-slur profanity is used
      strictly as a positive
      intensifier, self-directed frustration, or neutral reaction to an
      unexpected event, or
      when recounting third-person conflict or fictional quotes without
      endorsement; do not
      apply when profanity expresses hostility toward entertainers,
      athletes, or individuals, or
      uses gendered labels for women.
  Tier 2 - Diagnostic Inclusion Criteria (Check SECOND if no Tier 1
    override in S2 applies):
    - [S2.C1] -> OFFENSIVE: Assign OFFENSIVE when content directs
      second-person insults, hostile
      profanity, or angry name-calling at specific individuals,
      entertainers, or athletes, or
      features explicit sexual vulgarity and first-person taboo or
      disaster shock humor without
      protected-group bias; do not apply to non-profane political or
      sports criticism, fictional
      quotes, or third-person everyday narratives.
\end{verbatim}}

\subsection{Unfair-ToS (LexGLUE) Frozen Official Policy Prompt and Learned LEDGER Playbook}

\paragraph{Frozen Official Policy Prompt ($P_0$, EU Council Directive 93/13/EEC / CLAUDETTE Eight-Category Taxonomy).}
{\footnotesize\begin{verbatim}
You are a legal compliance classifier enforcing European Union consumer
  protection policy (EU Council Directive 93/13/EEC on Unfair Terms in
  Consumer Contracts, as operationalized in the CLAUDETTE / LexGLUE
  Unfair-ToS standard). Every Terms of Service (ToS) clause must be
  assigned to exactly one policy category: SUBSTANTIVE_UNFAIR,
  PROCEDURAL_UNFAIR, or COMPLIANT.

Section S1 - Substantive Unfair Service & Liability Terms
  (SUBSTANTIVE_UNFAIR):
Assign SUBSTANTIVE_UNFAIR to contractual clauses that create a significant
  imbalance in substantive rights and obligations to the detriment of the
  consumer under four statutory categories:
  (1) Limitation of liability: excluding or limiting the provider's legal
      liability for service failures, data loss, security breaches,
      third-party actions, or damages.
  (2) Unilateral termination: allowing the provider to terminate, suspend,
      or restrict a user's account or service access at its sole discretion
      or without specific objective grounds.
  (3) Unilateral change: allowing the provider to unilaterally modify,
      amend, or discontinue contract terms or service features.
  (4) Content removal: allowing the provider to delete, remove, or refuse
      user content at its sole discretion or without prior notice.

Section S2 - Procedural, Forum, Governing-Law & Assent Unfair Terms
  (PROCEDURAL_UNFAIR):
Assign PROCEDURAL_UNFAIR to contractual clauses that restrict a consumer's
  procedural remedies, legal forum, or informed consent under four
  statutory categories (when Section S1 does not apply):
  (5) Arbitration: requiring binding or mandatory arbitration or waiving
      class/representative actions.
  (6) Jurisdiction: requiring disputes to be litigated exclusively in a
      specific court or jurisdiction away from the consumer's residence.
  (7) Choice of law: specifying that the contract is governed by a
      designated national or state law without expressly preserving
      mandatory consumer protection laws of the consumer's residence.
  (8) Contract by using: binding the user to the Terms of Service merely by
      browsing, accessing, installing, or using the service without
      explicit affirmative agreement.

Section S3 - Fair & Compliant Contractual Clauses (COMPLIANT):
Assign COMPLIANT to contractual clauses that do not violate Section S1 or
  Section S2, including user obligations, intellectual property notices,
  user-initiated cancellation procedures, third-party billing descriptions,
  or balanced provisions that respect consumer statutory rights.
\end{verbatim}}

\paragraph{Learned Two-Tier Decision Playbook ($\mathcal{M}_{\text{opt}} = \text{Gemini 4 Argon}$, $86.01\%$ Test Macro-F1, $86.35\%$ Binary Unfairness F1, $9.79\%$ Subgroup FPR).}
{\footnotesize\begin{verbatim}
### DETAILED POLICY INTERPRETATION PLAYBOOK (STRICT TWO-TIER PRECEDENCE)
This Playbook provides the detailed operational interpretation of the
  Official Policy above. Apply it sequentially:
1. Substantive Unfair Service & Liability Boundary (Section S1): Whenever a
  contract clause addresses liability, warranties, unilateral
  termination/suspension, unilateral modification, or content removal,
  evaluate Section S1 FIRST. Check S1 Tier 1 Safe-Harbor Clauses FIRST;
  apply an S1 Tier 1 clause ONLY when its contractual safe-harbor condition
  holds AND none of its 'do not apply when' disqualifiers are triggered;
  otherwise evaluate S1 Tier 2 Substantive Unfairness Inclusion Clauses.
2. Procedural, Forum, Governing-Law & Assent Boundary (Section S2): Whenever
  SUBSTANTIVE_UNFAIR does not apply, evaluate Section S2 to distinguish
  PROCEDURAL_UNFAIR from COMPLIANT. Apply S2 Tier 1 (COMPLIANT) ONLY when
  dispute, law, or acceptance language respects consumer statutory
  protections AND none of its 'do not apply when' disqualifiers apply; if
  any S2 Tier 2 mandatory arbitration, exclusive distant jurisdiction,
  unqualified choice-of-law, or browsewrap assent clause applies, assign
  PROCEDURAL_UNFAIR; otherwise follow the Official Policy.

#### Section S1 - Substantive Unfair Service & Liability Boundary
  Tier 1 - Safe-Harbor Overrides (Check FIRST; overrides Tier 2 in S1):
    - [S1.B1] -> COMPLIANT: Assign COMPLIANT when the clause grants the
      user the right to terminate their own account or states that mutual
      obligations survive termination; do not apply when the clause imposes
      a consumer liability cap or unilateral provider termination right.
    - [S1.B2] -> COMPLIANT: Assign COMPLIANT when the clause clarifies the
      scope of the service or states the user is responsible for their own
      interactions without shifting statutory liability; do not apply when
      the clause explicitly disclaims the provider's legal liability for
      third-party actions or service failures.
    - [S1.B3] -> PROCEDURAL_UNFAIR: Assign PROCEDURAL_UNFAIR when the
      clause explicitly states that continued use of the service after a
      contract modification or notice period constitutes acceptance of the
      revised terms; do not apply when the clause instructs the user to
      review terms periodically or to discontinue use if they disagree.
  Tier 2 - Diagnostic Inclusion Criteria (Check SECOND in S1):
    - [S1.C1] -> SUBSTANTIVE_UNFAIR: Assign SUBSTANTIVE_UNFAIR when the
      clause excludes, caps, or limits the provider's legal liability for
      damages, service failures, data loss, or breaches; do not apply when
      the clause merely disclaims warranties of accuracy or uninterrupted
      availability without excluding damages liability.
    - [S1.C2] -> SUBSTANTIVE_UNFAIR: Assign SUBSTANTIVE_UNFAIR when the
      clause allows the provider to unilaterally suspend or terminate user
      access at any time or sole discretion; do not apply when the clause
      is a section heading or cancels only specific prohibited transactions.
    - [S1.C3] -> SUBSTANTIVE_UNFAIR: Assign SUBSTANTIVE_UNFAIR when the
      clause allows the provider to unilaterally amend terms or pricing, or
      obligates the user to periodically review the agreement for changes;
      do not apply when the clause merely explains reasons for updates or
      advises users to cancel if they reject updates.
    - [S1.C4] -> SUBSTANTIVE_UNFAIR: Assign SUBSTANTIVE_UNFAIR when the
      clause allows the provider at its discretion or at any time to
      remove, block, reclaim, or delete user content or identifiers; do not
      apply when the clause sets user eligibility or conduct obligations.

#### Section S2 - Procedural, Forum, Governing-Law & Assent Boundary
  Tier 1 - Safe-Harbor Overrides (Check FIRST; overrides Tier 2 in S2):
    - [S2.B1] -> COMPLIANT: Assign COMPLIANT when the clause details
      ancillary dispute resolution rules, including arbitration opt-out
      procedures, addresses for legal notices, exceptions for small claims,
      or severability of unenforceable terms; do not apply when the clause
      contains the core mandate requiring binding arbitration.
  Tier 2 - Diagnostic Inclusion Criteria (Check SECOND in S2):
    - [S2.C1] -> PROCEDURAL_UNFAIR: Assign PROCEDURAL_UNFAIR when the
      clause contains the primary agreement binding the user and provider
      to resolve disputes through mandatory individual arbitration and
      waive class actions; do not apply when the clause is a section title
      or explains survival or institution rules.
    - [S2.C2] -> PROCEDURAL_UNFAIR: Assign PROCEDURAL_UNFAIR when the
      clause mandates exclusive court litigation in a company-designated
      jurisdiction or designates a governing national or state law without
      preserving mandatory home-country consumer protections; do not apply
      when the clause specifies non-exclusive courts.
    - [S2.C3] -> PROCEDURAL_UNFAIR: Assign PROCEDURAL_UNFAIR when the
      clause is a browsewrap provision stating that merely accessing,
      browsing, downloading, or using the service binds the user to the
      Terms of Service; do not apply when the clause requires explicit
      affirmative clickwrap consent.
\end{verbatim}}

\section{Ablation Studies and Comparison with Published Supervised Models}
\label{app:ablations}

\subsection{Component and Baseline Ablation Study}
To isolate the contribution of each component in \textsc{LEDGER}, we evaluate four sets of ablations under identical supervision parity (receiving the exact same training examples, baseline errors $\mathcal{E}_{0}(\mathcal{D}_{\text{train}})$, subgroup tags, and annotator notes):
\begin{enumerate}
    \item \textbf{Naive Error-Only Playbook:} The Optimizer Model inspects the baseline training error set $\mathcal{E}_{0}(\mathcal{D}_{\text{train}})$ and generates a flat $9$--$11$ rule playbook $[R_1, \dots, R_K]$ without two-tier precedence, exclusion boundaries, or preservation-set verification.
    \item \textbf{Holistic Iterative Rewrite:} Modeling iterative whole-prompt optimization \cite{pryzant2023automatic,yang2024large,yuksekgonul2025optimizing}, the Optimizer Model rewrites the entire $10$--$12$ rule playbook across $T=3$ optimization rounds based on training confusion matrices, false positives, false negatives, and examples from the natural preservation set $\mathcal{P}_{0}(\mathcal{D}_{\text{train}})$, selecting the best round via the validation objective.
    \item \textbf{\textsc{LEDGER} (Round~1 Only vs.\ Full Lock-Repair-Prune):} Comparing the initial two-tier Decision Playbook from Round~1 against the final playbook after iterative Lock-Repair-Prune updates, across both $240/100$ and $480/200$ training/validation split sizes.
    \item \textbf{Subgroup Fairness Tolerance ($\delta_{\text{FPR}}$) Operating Points on SBIC.v2:} Comparing strict fairness ($\delta_{\text{FPR}} = 0.0$), balanced safe-harbor retention, and recall-focused optimization on SBIC.v2.
\end{enumerate}

\begin{table*}[h]
\centering
\caption{Component and baseline ablation study across \textbf{HateXplain} ($N_{\text{test}}=500$), \textbf{SBIC.v2} ($N_{\text{test}}=350$), and \textbf{Unfair-ToS} ($N_{\text{test}}=350$) using $\mathcal{M}_{\text{opt}} = \text{Gemini 4 Argon}$ and $\mathcal{M}_{\text{eval}} = \text{Gemini 3.1 Flash-Lite}$. Fixes ($W \to R$) and Regressions ($R \to W$) are measured relative to Policy Prompt Only.}
\label{tab:ablation_main}
\resizebox{\textwidth}{!}{%
\begin{tabular}{lllcccccc}
\toprule
\textbf{Benchmark} & \textbf{Ablation / Method Variant} & \textbf{Train/Val Split (Rules)} & \textbf{Acc (\%)} & \textbf{Macro-F1 (\%)} & \textbf{Fixes / Regs} & \textbf{Subgroup FPR (\%)} & \textbf{Cite (\%)} \\
\midrule
\textbf{HateXplain} & Policy Prompt Only ($\emptyset$) & Zero ($0$ rules) & 60.20 & 58.88 & $0 / 0$ & 40.88 & 0.0 \\
 & Naive Error-Only Playbook (Flat Rules) & $480/200$ ($10$ rules) & 58.80 & 59.11 & $89 / 96$ & 7.30 & 0.0 \\
 & Holistic Iterative Rewrite ($T=3$) & $480/200$ ($10$ rules) & 68.40 & 67.10 & $68 / 27$ & 27.74 & 0.0 \\
 & \textsc{LEDGER} (Round 1 Only, w/o Lock-Repair-Prune) & $240/100$ ($11$ rules) & 61.40 & 60.48 & $56 / 50$ & 35.77 & 99.6 \\
 & \textsc{LEDGER} (Full, Smaller Split) & $240/100$ ($11$ rules) & 65.80 & 65.36 & $62 / 34$ & 21.17 & \textbf{100.0} \\
 & \textbf{\textsc{LEDGER} (Full)} & $480/200$ ($10$ rules) & \textbf{68.60} & \textbf{68.35} & $79 / 37$ & \textbf{16.79} & 99.6 \\
\midrule
\textbf{SBIC.v2} & Policy Prompt Only ($\emptyset$) & Zero ($0$ rules) & 73.14 & 69.49 & $0 / 0$ & \textbf{7.41} & 0.0 \\
 & Naive Error-Only Playbook (Flat Rules) & $480/200$ ($9$ rules) & 77.43 & 64.54 & $50 / 35$ & 44.44 & 0.0 \\
 & Holistic Iterative Rewrite ($T=3$) & $480/200$ ($10$ rules) & 75.43 & 70.61 & $31 / 23$ & 33.33 & 0.0 \\
 & \textsc{LEDGER}-Strict ($\delta_{\text{FPR}}=0.0$) & $480/200$ ($10$ rules) & 75.43 & 68.96 & $37 / 29$ & 14.81 & \textbf{98.9} \\
 & \textsc{LEDGER}-Flow (Round 1 Only) & $480/200$ ($9$ rules) & 78.00 & 70.65 & $46 / 29$ & 29.63 & 93.1 \\
 & \textsc{LEDGER}-Flow (Full, Group-Targeting F1 $= 85.81\%$) & $480/200$ ($8$ rules) & 79.43 & 72.49 & $44 / 22$ & 37.04 & 96.3 \\
 & \textbf{\textsc{LEDGER} (Full Balanced)} & $240/100$ ($10$ rules) & \textbf{80.00} & \textbf{73.65} & $49 / 25$ & 18.52 & 96.6 \\
\midrule
\textbf{Unfair-ToS} & Policy Prompt Only ($\emptyset$) & Zero ($0$ rules) & 75.14 & 74.89 & $0 / 0$ & 19.15 & 0.0 \\
 & \textsc{LEDGER} (Round 1 Only, w/o Lock-Repair-Prune) & $480/200$ ($11$ rules) & 83.43 & 83.80 & $37 / 8$ & 12.34 & \textbf{95.4} \\
 & \textbf{\textsc{LEDGER} (Full)} & $480/200$ ($11$ rules) & \textbf{85.71} & \textbf{86.01} & $46 / 9$ & \textbf{9.79} & 88.0 \\
\bottomrule
\end{tabular}%
}
\end{table*}

\paragraph{Effect of Two-Tier Precedence vs.\ Flat Error Rules.}
Table~\ref{tab:ablation_main} illustrates the Rule-Overlap Trap when rules lack two-tier precedence and exclusion boundaries. On HateXplain, the Naive Error-Only Playbook fixes $89$ baseline errors but causes $96$ collateral regressions on $\mathcal{P}_{0}(\mathcal{D}_{\text{test}})$, reducing overall accuracy from $60.20\%$ to $58.80\%$ and collapsing \texttt{HATESPEECH} F1 from $72.04\%$ to $57.74\%$. On SBIC.v2, Naive Error-Only over-corrects in the opposite direction by escalating ambiguous posts to \texttt{HATESPEECH}, collapsing \texttt{OFFENSIVE} F1 from $51.06\%$ to $26.32\%$, reducing three-class Macro-F1 from $69.49\%$ to $64.54\%$, and raising Subgroup Hate-FPR from $7.41\%$ to $44.44\%$. Across all six $(\mathcal{M}_{\text{opt}}, \mathcal{M}_{\text{eval}})$ pairs on HateXplain, Naive Error-Only causes $57$--$110$ collateral regressions ($59.11\%$--$66.62\%$ Macro-F1) and Holistic Iterative Rewrite causes $27$--$59$ regressions ($62.16\%$--$67.10\%$ Macro-F1), whereas \textsc{LEDGER} limits regressions to $28$--$44$ while providing $95.2\%$--$99.6\%$ explicit clause attribution.

\paragraph{Effect of Lock-Repair-Prune Refinement Over Round~1.}
Comparing \textsc{LEDGER} (Round~1 Only) against \textsc{LEDGER} (Full) in Table~\ref{tab:ablation_main} isolates the contribution of clause-level Lock-Repair-Prune updates. On HateXplain ($240/100$), Round~2 Lock-Repair-Prune locks $4$ verified clauses and repairs the remaining clauses, improving held-out Macro-F1 by $+4.88$ points ($60.48\%$ to $65.36\%$), reducing collateral regressions from $50$ to $34$, and lowering Subgroup Hate-FPR by $14.60$ percentage points ($35.77\%$ to $21.17\%$). On SBIC.v2 ($480/200$), Lock-Repair-Prune locks $3$ clauses, prunes $1$ inactive clause, and repairs $5$ clauses, raising three-class Macro-F1 by $+1.84$ points ($70.65\%$ to $72.49\%$) and binary \emph{Targeting Group?} F1 to $85.81\%$. On Unfair-ToS ($480/200$), Lock-Repair-Prune locks $6$ verified clauses (\texttt{S1.B2}, \texttt{S1.B3}, \texttt{S1.C3}, \texttt{S2.B1}, \texttt{S2.C2}, and \texttt{S2.C3}) and repairs $5$ clauses, raising three-section Macro-F1 by $+2.21$ points ($83.80\%$ to $86.01\%$) and lowering Subgroup FPR from $12.34\%$ to $9.79\%$.

\subsection{Contextual Comparison with Published Supervised Models}
For contextual comparison against weight-updated models trained on the full benchmark corpora, Table~\ref{tab:published_ref} summarizes the headline metrics reported by published supervised models alongside Policy Prompt Only and \textsc{LEDGER}. Because the published supervised models update model weights on the full training splits ($\sim 15\text{K}$--$35\text{K}$ examples) and evaluate on the full test corpora (including the $\sim 10.7\%$ unfair-clause prior on Unfair-ToS), they are presented as contextual references rather than direct split-matched comparisons.

\begin{table*}[h]
\centering
\caption{Contextual comparison between published full-corpus supervised models and zero-weight-update playbook enforcement (\textbf{Policy Prompt Only} and \textbf{\textsc{LEDGER}} with $\mathcal{M}_{\text{eval}} = \text{Gemini 3.1 Flash-Lite}$).}
\label{tab:published_ref}
\resizebox{\textwidth}{!}{%
\begin{tabular}{llllc}
\toprule
\textbf{Benchmark} & \textbf{Model / Method} & \textbf{Training Paradigm \& Supervision} & \textbf{Evaluation Metric} & \textbf{Score (\%)} \\
\midrule
\textbf{HateXplain} & Standard BERT \cite{mathew2021hatexplain} & Full Weight Fine-Tuning ($\sim 15\text{K}$ posts, Full Test Ref.) & 3-Class Macro-F1 & 67.40 \\
 & BERT-HateXplain (+Rationales) \cite{mathew2021hatexplain} & Full Weight Fine-Tuning + Attention Span Supervision (Ref.) & 3-Class Macro-F1 & 68.70 \\
 & Google Jigsaw Perspective API \cite{lees2022perspective} & Production Supervised Toxicity Classifier (Ref.) & 3-Class Macro-F1 & 70.30 \\
 & Policy Prompt Only ($\emptyset$) & Zero Weight Updates ($0$ rules, $N_{\text{test}}=500$) & 3-Class Macro-F1 & 58.88 \\
 & \textbf{\textsc{LEDGER} (Ours)} & \textbf{Zero Weight Updates ($480$ train examples, $10$ rules, $N_{\text{test}}=500$)} & \textbf{3-Class Macro-F1} & \textbf{68.35} \\
\midrule
\textbf{SBIC.v2} & \citet{sap2020social} Transformer & Full Weight Fine-Tuning on SBIC Corpus (Full Test Ref.) & Offensiveness / Group-Target F1 & $78.80 / 69.90$ \\
 & Policy Prompt Only ($\emptyset$) & Zero Weight Updates ($0$ rules, $N_{\text{test}}=350$) & Offensiveness / Group-Target F1 & $83.92 / 78.60$ \\
 & \textbf{\textsc{LEDGER} (Ours)} & \textbf{Zero Weight Updates ($240$ train examples, $10$ rules, $N_{\text{test}}=350$)} & \textbf{Offensiveness / Group-Target F1} & $\mathbf{85.48 / 83.92}$ \\
\midrule
\textbf{Unfair-ToS} & Linear SVM \cite{lippi2019claudette} & Supervised SVM on CLAUDETTE Corpus (Full Test Ref.) & Binary Unfairness F1 & 75.20 \\
 & Legal-BERT \cite{chalkidis2022lexglue} & Domain-Pretrained + Full Weight Fine-Tuning (Full Test Ref.) & Binary Unfairness F1 & 81.3--83.0 \\
 & Policy Prompt Only ($\emptyset$) & Zero Weight Updates ($0$ rules, Stratified $N_{\text{test}}=350$) & Binary Unfairness F1 & 79.61 \\
 & \textbf{\textsc{LEDGER} (Ours)} & \textbf{Zero Weight Updates ($480$ train examples, $11$ rules, Stratified $N_{\text{test}}=350$)} & \textbf{Binary Unfairness F1} & \textbf{86.35} \\
\bottomrule
\end{tabular}%
}
\end{table*}

\end{document}